\documentclass[sigconf,authorversion,nonacm]{acmart}

\usepackage[utf8]{inputenc}

\usepackage{times}
\usepackage{latexsym}
\usepackage{amsmath}
\usepackage{amsfonts}
\usepackage{graphicx}
\usepackage{multirow}
\usepackage{verbatim}
\usepackage[justification=centering]{subfig}
\usepackage{tikz}
\usepackage{hyperref}
\usetikzlibrary{fit,positioning}
\usepackage{comment}
\usepackage{float}
\usepackage{cleveref}
\usepackage{amsthm}
\usepackage[linesnumbered,ruled,vlined]{algorithm2e}
\usepackage{multicol}
\usepackage{xcolor}

\usepackage[utf8]{inputenc}

\setcopyright{none}
\begin{document}


\title{Subject Granular Differential Privacy in Federated Learning}

\author{Virendra J. Marathe}
\email{virendra.marathe@oracle.com}
\affiliation{%
  \institution{Oracle Labs}
  \city{}
  \country{}
}

\author{Pallika Kanani}
\email{pallika.kanani@oracle.com}
\affiliation{%
  \institution{Oracle Labs}
  \city{}
  \country{}
}

\author{Daniel Peterson}
\email{daniel.peterson@oracle.com}
\affiliation{%
  \institution{Oracle Labs}
  \city{}
  \country{}
}

\author{Guy Steele Jr.}
\email{guy.steele@oracle.com}
\affiliation{%
  \institution{Oracle Labs}
  \city{}
  \country{}
}

\begin{abstract}
  This paper considers \emph{subject level privacy} in the FL setting,
  where a subject is an individual whose private information is
  embodied by several data items either confined within a single
  federation user or distributed across multiple federation users.
  We propose two new algorithms that enforce subject level DP at each
  federation user \emph{locally}.  Our first algorithm, called
  \localgroupdp, is a straightforward application of \emph{group
    differential privacy} in the popular DP-SGD algorithm.  Our second
  algorithm is based on a novel idea of \emph{hierarchical gradient
    averaging (\higradavg)} for subjects participating in a training
  mini-batch.  We also show that user level \emph{Local Differential
    Privacy (LDP)} naturally guarantees subject level DP.  We 
  observe the problem of \emph{horizontal composition} of subject
  level privacy loss in FL -- subject level privacy loss incurred at
  individual users composes across the federation.
  We formally prove the subject level DP guarantee for our algorithms,
  and also show their effect on model utility loss.  Our empirical
  evaluation on FEMNIST and Shakespeare datasets shows that
  \localgroupdp\ delivers the best performance among our algorithms.
  However, its model utility lags behind that of models trained using
  a DP-SGD based algorithm that provides a \emph{weaker} item level
  privacy guarantee.  Privacy loss amplification due to subject
  sampling fractions and horizontal composition remain key
  challenges for model utility.

\end{abstract}

\newcommand{\sref}[2]{\hyperref[#2]{#1 \ref*{#2}}}

\newcommand{\dpsgd}{\emph{DP-SGD}}
\newcommand{\dpfedsgd}{\emph{DP-FedSGD}}
\newcommand{\localdpsgd}{\emph{LocalItemDP}}
\newcommand{\usercentraldp}{\emph{CentralUserDP}}
\newcommand{\userlocaldpsgd}{\emph{UserLDP}}
\newcommand{\userlocalsgd}{\emph{User-Local-SGD}}
\newcommand{\userlocaloutperturb}{\emph{User-Local-Out-Perturb}}
\newcommand{\fedsgd}{\emph{FedSGD}}
\newcommand{\fedavg}{\emph{FedAvg}}
\newcommand{\individual}{\emph{Individual}}
\newcommand{\localgroupdp}{\emph{LocalGroupDP}}
\newcommand{\higradavg}{\emph{HiGradAvgDP}}
\newcommand{\centralsubdp}{\emph{CentralSubDP}}
\newcommand{\localsubdp}{\emph{LocalSubDP}}

\newcommand{\femnistlarge}{\emph{FEMNIST-Large}}
\newcommand{\femnistsmall}{\emph{FEMNIST-Small}}


\SetKwInput{KwInput}{Parameters}
\SetKwInput{KwOutput}{Output}

\maketitle

\thispagestyle{plain}
\pagestyle{plain}

\section{Introduction}
\label{sec:intro}

Data privacy enforcement, using Differential Privacy
(DP)~\cite{dwork06,dwork14}, in the Federated Learning (FL)
setting~\cite{konecny15} has been explored at two granularities: (i)
\emph{item level privacy}, where use of each data item in model
training is obfuscated~\cite{abadi16}; and (ii) \emph{user level
  privacy}, where participation of each federation user is
hidden~\cite{mcmahan18}.

Another dimension of DP in FL relates to the \emph{locale} (physical
location) of DP enforcement.  The most common alternatives are (i)
\emph{locally} at federation users~\cite{abadi16,dp-apple17,truex20},
and (ii) \emph{centrally} at the federation server~\cite{mcmahan18}.
The privacy enforcement locale is dictated by assumptions made about
the \emph{trust model} between federation users and the server.  A
trusted server enables central enforcement of DP at the server,
whereas the users may prefer to locally enforce DP with an untrusted
server.

In this paper, we assume that the federation users and the server
behave as \emph{honest-but-curious} participants in the federation:
They do not interfere with or manipulate the distributed training
protocol, but may be interested in analyzing received model
updates. Federation users do not trust each other or the federation
server, and must locally enforce privacy guarantees for their private
data.

User level privacy is perhaps the right privacy granularity in the
original \emph{cross-device} FL setting consisting millions of hand
held devices~\cite{bonawitz19,konecny15}.  Furthermore, central
enforcement of user level privacy~\cite{mcmahan18} appears to be the
most viable approach in that setting.  However, the \emph{cross-silo}
FL setting~\cite{kairouz19}, where federation users are organizations
that are themselves gatekeepers of data items of numerous individuals
(which we call ``subjects'' henceforth), offer much richer mappings
between subjects and their personal data.

In the simplest of use cases, a subject is embodied by a single data
item across the entire federation.  Thus item level privacy is
sufficient to guarantee subject level privacy.  However, many real
world use cases exhibit more complex subject to data mappings.
Consider a patient $P$ visiting different hospitals for treatment of
different ailments. Each hospital contains multiple data records
forming $P$’s health history. These hospitals may decide to
participate in a federation that uses their respective patients’
health history records in their training datasets. Thus distinct
health history records of the same data subject (e.g. $P$) can appear
in the datasets of multiple hospitals. In the end, it is the privacy
of these subjects that we want to preserve in a FL federation.

Item level privacy, irrespective of its enforcement locale, does not
suffice to protect privacy of $P$'s data.  That is because item level
privacy simply obfuscates participation of individual data items in
the training process~\cite{abadi16,dwork06,dwork14}.  Since a subject
may have multiple data items in the dataset, item level private
training may still leak a subject's data
distribution~\cite{liu20,mcmahan18}.  User level privacy enforced
centrally~\cite{mcmahan18} does not protect the privacy of $P$'s data
either.  User level privacy obfuscates each user's participation in
training~\cite{mcmahan18}.  However, a subject's data can be
distributed among several users, and it can be leaked when aggregated
through FL. In the worst case, multiple federation users may host only
the data of a single subject.  Thus $P$'s data distribution can be
leaked even if individual users' participation is obfuscated
centrally.

In this paper, we consider a third granularity of privacy --
\emph{subject level privacy}~\cite{wang21}\footnote{ Wang et
  al.~\cite{wang21} identify what we call subjects in this paper as
  users in their paper.}, where a subject is an individual whose
private data is spread across multiple data items, which can
themselves be distributed across multiple federation users.  The
notion of subject level privacy is not new, and in fact appears in
some of the original work on DP~\cite{dwork06,dwork14}.  However, most
existing work has either assumed a $1$-to-$1$ mapping between subjects
and data items~\cite{abadi16}, or has treated subjects as individual
silos of data (a.k.a.~users) in a collaborative learning setting such
as FL~\cite{mcmahan18}.  Recent work has addressed subject level
privacy in a centralized setting~\cite{levy21}, but no prior work has
addressed the problem in a distributed collaborative learning setting
such as FL.  To the best of our knowledge, our work is the first study
of subject level privacy in FL.

We formulate subject level privacy in terms of the classic definition
of differential privacy~\cite{dwork06}.
We present two novel algorithms -- \localgroupdp\ and \higradavg --
that achieve subject level DP in the FL setting.
We formally prove our algorithms' subject level DP guarantee.  We also
show that user level \emph{Local Differential Privacy (LDP)}, called
\userlocaldpsgd, provides the subject level DP guarantee.

We observe that subject level privacy loss at individual federation
users composes across users in the federation. We call this
\emph{horizontal composition}.  We show that, in the worst case,
horizontal composition is equivalent to composition of privacy loss in
iterative computations such as ML model training over mini-batches.
Consequently, the recent advances in adaptive composition
results~\cite{abadi16,dong19,dwork10,mironov17} apply to horizontal
composition.  This adds additional constraints on model training
either in terms of additional noise injection or in terms of the
amount of training permitted -- reduction in training rounds by a
factor of $\frac{1}{\sqrt{s}}$, where $s$ is the number of users
sampled in a training round.  These constraints adversely affect model
utility.

We formally analyze utility loss of models trained with our algorithms
in terms of excess population loss~\cite{bassily19,bassily14},
assuming $L$-Lipschitz convex loss functions.  We show that, compared
to the utility loss incurred by the item level DP enforcement
algorithm by Abadi et al.~\cite{abadi16} (\localdpsgd), utility loss
of models trained using \localgroupdp, \userlocaldpsgd, and
\higradavg\ is affected significantly by different factors.  In case
of \localgroupdp\ and \userlocaldpsgd, the utility degradation is
amplified by a quadractic factor of the group size per mini-batch (the
group size for \userlocaldpsgd\ is the size of the mini-batch).  For
\higradavg, the utility degradation is amplified by a quadratic factor
of the cardinality of the most frequently occuring subject in a
federation user's dataset.

Our empirical evaluation results, using the FEMNIST and Shakespeare
datasets~\cite{caldas18}, reflect our formal analysis of utility loss:
\localgroupdp\ and \userlocaldpsgd\ incur significant model utility
overheads (degradation over \localdpsgd\ of $15\%$ on FEMNIST and
$18\%$ on Shakespeare with \localgroupdp, and far worse with
\userlocaldpsgd).  \higradavg\ leads to model degradation over
\localdpsgd\ of $22\%$ on FEMNIST, and much worse utility than even
\userlocaldpsgd\ on Shakespeare.  This leaves us with the open problem
of building high utility algorithms that guarantee subject level DP in
FL.

The rest of the paper is organized as follows: Relevant definitions
appear in~\sref{Section}{sec:definition}.  Our algorithms and
horizontal composition are described in~\sref{Section}{sec:algos}; we
also prove their subject level DP
guarantee.~\sref{Section}{sec:utility-loss} formally shows the utility
loss incurred more generally by subject level DP enforcement, and
specifically by our algorithms.  Our empirical evaluation appears
in~\sref{Section}{sec:evaluation}, followed by conclusion
in~\sref{Section}{sec:conclusion}.

\section{Subject Level Differential Privacy}
\label{sec:definition}

We begin with the definition of Differential Privacy~\cite{dwork06}.  Informally, DP
bounds the maximum impact a single data item can have on the output of
a randomized algorithm $\mathcal{A}$.  Formally,

\begin{definition}
A randomized algorithm $\mathcal{A}: \mathcal{D} \rightarrow
\mathcal{R}$ is said to be ($\varepsilon$,$\delta$)-differentially
private if for any two \emph{adjacent} datasets $D$, $D' \in
\mathcal{D}$, and set $S \subseteq \mathcal{R}$,
\begin{eqnarray}
  \mathrm{Pr}[\mathcal{A}(D) \in S] \le
  e^{\varepsilon} \mathrm{Pr}[\mathcal{A}(D') \in S] + \delta
\label{eq:dp}
\end{eqnarray}
where $D$, $D'$ are adjacent to each other if they differ from each
other by a single data item.  $\delta$ is the probability of failure
to enforce the $\varepsilon$ privacy loss bound.
\end{definition}


The above definition provides item level privacy.  McMahan et
al.~\cite{mcmahan18} present an alternate definition for user level DP
in the FL setting.  Let $U$ be the set of $n$ users participating in a
federation, and $\mathcal{D}_i$ be the dataset of user $u_i \in
U$. Let $\mathcal{D}_U = \bigcup_{i=1}^n \mathcal{D}_i$.  Let
$\mathcal{R}$ be the range of models resulting from the FL training
process.

\begin{definition}
  Given a FL training algorithm $\mathcal{A}: \mathcal{D_U}
  \rightarrow \mathcal{R}$, we say that $\mathcal{A}$ is \emph{user
    level ($\varepsilon,\delta$)-differentially private} if for any
  two adjacent user sets $U$, $U' \subseteq \mathcal{U}$, and $S
  \subseteq \mathcal{R}$,
  \begin{eqnarray}
    \mathrm{Pr}[\mathcal{A}(\mathcal{D}_U) \in S] \le
    e^{\varepsilon} \mathrm{Pr}[\mathcal{A}(\mathcal{D}_{U'}) \in S] + \delta
    \label{eq:user-dp}
  \end{eqnarray}
  where $U$, $U'$ are adjacent user sets differing by a single user.
  \label{def-user-level-dp}
\end{definition}

Let $\mathcal{Q}$ be the set of subjects whose data is hosted by the
federation's users $U$.  Our definition of subject level DP
is based on the observation that, even though the data of individual
subjects $s \in \mathcal{Q}$ may be physically scattered across
multiple users in $U$, the aggregate data across
$U$ can be logically divided into its subjects in
$\mathcal{Q}$ (i.e.~$\mathcal{D}_U = \bigcup_{s \in \mathcal{Q}}
\mathcal{D}_s$).

\begin{definition}
  Given a FL training algorithm $\mathcal{A}:\mathcal{D_U} \rightarrow
  \mathcal{R}$, we say that $\mathcal{A}$ is \emph{subject level
    ($\varepsilon,\delta$)-differentially private} if for any two
  adjacent subject sets $Q$, $Q' \subseteq \mathcal{Q}$, and $S
  \subseteq \mathcal{R}$,
  \begin{eqnarray}
    \mathrm{Pr}[\mathcal{A}(\mathcal{D}_Q) \in S] \le
    e^{\varepsilon} \mathrm{Pr}[\mathcal{A}(\mathcal{D}_{Q'}) \in S] +
    \delta
    \label{eq:subject-dp}
  \end{eqnarray}
  where $Q$ and $Q'$ are adjacent subject sets if they differ from
  each other by at most a single subject.
\end{definition}

Note that the above definition completely ignores the notion of users
in a federation.  This user obliviousness is crucial to make the
definition work for both cases: (i) where a subject's data items are
confined to a single user (e.g. for cross-device FL settings), and
(ii) where a subject's data items are spread across multiple users
(e.g. for cross-silo FL settings)~\cite{wang21}.

\section{Enforcing Subject Level Differential Privacy}
\label{sec:algos}

We assume a federation that contains a federation server that is
responsible for (i) initialization and distribution of the model
architecture to the federation users, (ii) coordination of training
rounds, (iii) aggregation and application of model updates coming from
different users in each training round, and (iv) redistribution of the
updated model back to the users.  Each federation user (i) receives
updated models from the federation server, (ii) retrains the received
models using its private training data, and (iii) returns updated
model parameters to the federation server.

Our algorithms enforce subject level DP \emph{locally} at each user.
But to prove the privacy guarantee for any subject, across the entire
federation, we must ensure that the local subject level DP guarantee
\emph{composes} correctly through the \emph{global} aggregation, at
the federation server, of parameter updates received from these users.
To that end we break down the federated training round into two
functions: (i) $\mathcal{F}_l$, the user's training algorithm that
enforces subject level DP locally, and (ii) $\mathcal{F}_g$, the
server's operation that aggregates parameter updates received from all
the users.  We first present our algorithms for $\mathcal{F}_l$ and
show that they locally enforce subject level DP.  Thereafter we show
how an instance of $\mathcal{F}_g$ that simply averages parameter
updates (at the federation server) composes the subject level DP
guarantee across multiple users in the federation.

Our algorithms are based on a federated version of the~\dpsgd\
algorithm by Abadi et al.~\cite{abadi16}.  \dpsgd\ was originially not
designed for FL, but can be easily extended to enforce item level DP
in FL: The federation server samples a random set of users for each
training round and sends them a request to perform local training.
Each user trains the model locally using \dpsgd.
Formally, the parameter update at step $t$ in \dpsgd\ using a
mini-batch of size $b$ can be summarized in the following equation:
\begin{equation}
  \Theta_t = \Theta_{t-1} + \frac{\eta}{b} (\sum_{i=1}^b
  \mathcal{\triangledown L}_i^C(\Theta_{t-1}) + \mathcal{N}(0, C^2
  \sigma^2))
  \label{eq:dpsgd}
\end{equation}

\noindent where, $\mathcal{\triangledown L}_i^C$ is the loss
function's gradient, for data item $i$ in the mini-batch, clipped by
the norm threshold of $C$, $\sigma$ is the noise scale calculated
using the moments accountant method, $\mathcal{N}$ is the Gaussian
distribution used to calculate noise, and $\eta$ is the learning rate.
Note that the gradient for each data item is clipped separately to
limit the influence (\emph{sensitivity}) of each data item in the
mini-batch.  The $\sigma$ is derived from Theorem 1 in~\cite{abadi16}

\begin{theorem}
  There exist constants $c_1$ and $c_2$ such that given the sampling
  probability $q = \frac{B}{|D|}$, where $B$ is the mini-batch size,
  $D$ the training dataset, and $T$ is the number of steps, for any
  $\varepsilon < c_1 q^2 T$, \dpsgd\ enforces item level
  ($\varepsilon$,$\delta$)-differential privacy for any $\delta > 0$
  if we choose
  \begin{align*}
    \sigma \ge c_2 \frac{q\sqrt{T log(1/\delta)}}{\varepsilon}
  \end{align*}
\label{thm:dpsgd}
\end{theorem}

The users ship back updated model parameters to the federation server,
which averages the updates received from all the sampled users.  The
server redistributes the updated model and triggers another training
round if needed.  The original paper~\cite{abadi16} also proposed the
moments accountant method for tighter composition of privacy loss
bounds compared to prior work on strong composition~\cite{dwork10}.
We call this described algorithm \localdpsgd.

\subsection{Locally Enforced Group Level Differential Privacy}

Intuitively, \localdpsgd\ enforces item level DP by injecting noise
proportional to any sampled data item's influence in each mini-batch.
In order to extend this approach to enforce subject level DP, we need
to precisely calibrate noise proportional to a data subject's
influence on a mini-batch's gradients.
A direct method to attain that is by obfuscating the effects of the
\emph{group} of data items belonging to the same subject.  We can
apply the formalism of \emph{group differential privacy} to achieve
this group level obfuscation.  The following theorem is a restatement,
in our notation, of Theorem $2.2$ and the associated footnote $1$ from
Dwork and Roth~\cite{dwork14}, §2.3:

\begin{theorem}
Any ($\varepsilon,\delta$)-differentially private randomized algorithm
$\mathcal{A}:\mathcal{D} \rightarrow \mathcal{R}$ is ($k \varepsilon,k
e^{(k-1)\varepsilon}\delta$)-differentially private for groups of size
$k \geq 2$.  That is, for all $D, D' \in \mathcal{D}$ such that $D$
and $D'$ differ in at most $k$ data items, and for all $S \subseteq
\mathcal{R}$,
\begin{eqnarray*}
\mathrm{Pr}[\mathcal{A}(D) \in S] & \leq & e^{k\varepsilon}
\mathrm{Pr}[\mathcal{A}(D') \in S] + k e^{(k-1)\varepsilon} \delta
\end{eqnarray*}
\label{thm:group-dp}
\end{theorem}

The proof of~\autoref{thm:group-dp} appears in the appendix. The
following definition is a direct consequence.

\begin{definition}
  We say that a randomized algorithm $\mathcal{A}$ is
  \emph{($\mathcal{E},\Delta$)-group differentially private} for a
  group size of $g$, if $\mathcal{A}$ is
  ($\epsilon$,$\delta$)-differentially private, where $\mathcal{E} = g
  \epsilon$, and $\Delta = g e^{(g-1)\epsilon}$.
\label{def:group-dp}
\end{definition}

Clearly, group DP incurs a big linear penalty on the privacy loss
$\varepsilon$, and an even bigger penalty in the failure probability
($g e^{(g-1) \varepsilon} \delta$).  Nevertheless, if $g$ is
restricted to a small value (e.g. $2$) the group DP penalty
may be acceptable.

In the FL setting, subject level DP immediately follows from group DP
for every sampled mini-batch of data items at every federation user.
Let $\mathcal{B}$ be a sampled mini-batch of data items at a user
$u_i$, and $\mathcal{R}$ be the domain space of the ML model being
trained in the FL setting.

\begin{theorem}
Let training algorithm $\mathcal{A}_g: \mathcal{B} \rightarrow
\mathcal{R}$ be group differentially private for groups of size $g$,
and $l$ be the largest number of data items belonging to any single
subject in $\mathcal{B}$.  If $l \leq g$, then $\mathcal{A}_g$ is
subject level differentially private.
\label{thm:group-sub-dp}
\end{theorem}

Composition of group DP guarantees over multiple mini-batches and
training rounds also follows established DP composition
results~\cite{abadi16,dwork10,mironov17}.  For instance, the moments
accountant method by Abadi et al.~\cite{abadi16} shows that given an
($\varepsilon$,$\delta$)-DP gradient computation for a single
mini-batch, the full training algorithm, which consists of $T$
mini-batches and a mini-batch sampling fraction of $q$, is $(O(q
\varepsilon \sqrt{T}),\delta)$-differentially
private. \sref{Theorem}{thm:group-dp} implies that the same algorithm
is $(O(g q \varepsilon \sqrt{T},g e^{(g-1) \varepsilon}
\delta)$-differentially private for a group of size $g$.

We now present our new FL training algorithm, \localgroupdp, that
guarantees group DP.
We make a critical assumption in \localgroupdp: Each user can
determine the subject for any of its data items.  Absent this
assumption, the user may need to make the worst case assumption that
all data items used to train the model belong to the same subject. On
the other hand, these algorithms are \emph{strictly local}, and do not
require that the identity of the subjects be resolved across users.

\begin{algorithm}[t]

  {
  \KwInput{Set of $n$ users $\mathcal{U} = {u_i,u_2,...,u_n}$;
    $\mathcal{D}_i$, the dataset of user $u_i$; $M$, the model to be
    trained; $\theta$, the parameters of model $M$ ; gradient norm
    bound $C$; sample of users $U_s$; mini-batch size $B$; $Z$,
    largest group size in a mini-batch, $\sigma_Z$, precomputed noise
    scale for group of size $Z$; $R$ training rounds; $T$ batches per
    round; the learning rate $\eta$.
  }

  \setlength{\columnsep}{1pt}

    \SetKwProg{Fn}{}{}{}
    \Fn{LocalGroupDP($u_i$):}{
      \For{$t = 1$ \KwTo $T$}{
        $\mathcal{B}$ = random sample of $B$ data items from $\mathcal{D}_i$\\
        \For{$s_i \in \mathcal{B}$}{ 
          {\bf Compute gradients:}\\
          $g(s_i) = \triangledown \mathcal{L}(\theta,s_i)$\\
          {\bf Clip gradients:}\\
          $\bar{g}(s_i) = Clip(g(s_i),C)$\\
        }
        $Z = LrgGrpCnt(\mathcal{B})$\\
        $\tilde{g}_s = \frac{1}{B} (\sum_i \bar{g}(s_i) + \mathcal{N}(0, \sigma_Z^2 C^2 {\bf I}))$\\
        $\theta = \theta - \eta \tilde{g}_s$\\
      }
      {\bf return} $M$\\
    }
    \SetKwProg{Fn}{}{}{}
    \Fn{Server Loop:}{
      \For{$r = 1$ \KwTo $R$}{
        $U_s$ = sample $s$ users from $\mathcal{U}$\\
        \For{$u_i \in U_s$}{
          $\theta_i = \emph{LocalGroupDP}(u_i$)\\
        }
        $\theta = \frac{1}{s} \sum_i \theta_i$ \\
        Send $M$ to all users in $\mathcal{U}$\\
      }
    }
    
  }
  \caption{Pseudo code for \localgroupdp\ that guarantees
    \emph{subject level} DP via group DP enforcement.}
\label{algo:local-group-dp}
\end{algorithm}

\localgroupdp\ (~\sref{Algorithm}{algo:local-group-dp}) enforces
subject level privacy locally at each user.  Like prior
work~\cite{abadi16,mcmahan18,song13}, we enforce DP in
\localgroupdp\ by adding carefully calibrated Gaussian noise in each
mini-batch's gradients.  Each user clips gradients for each data item
in a mini-batch to a clipping threshold $C$ prescribed by the
federation server.  The clipped gradients are subsequently averaged
over the mini-batch.  The clipping step bounds the \emph{sensitivity}
of each mini-batch's gradients to $C$.

To enforce group DP, \localgroupdp\ also locally tracks the item count
of the subject with the largest number of items in the sampled
mini-batch (\emph{LrgGrpCnt($\mathcal{B}$)} in
\sref{Algorithm}{algo:local-group-dp}).  This count determines the
\emph{group size} needed to enforce group DP for that mini-batch.
This group size, $Z$ in~\sref{Algorithm}{algo:local-group-dp}, helps
determine the noise scale $\sigma_Z$, given the target privacy
parameters $(\mathcal{E},\Delta)$ over the entire training round.
More specifically, we use the moments accountant method
and~\sref{Definition}{def:group-dp} to calculate $\sigma_Z$ for
$\varepsilon = \mathcal{E}/Z$, and $\delta = \Delta / (Z
e^{(Z-1)\frac{\mathcal{E}}{Z}})$.
$\sigma_Z$ is computed using the moments accountant method.  The rest
of the parameters to calculate $\sigma_Z$ --~$\mathcal{E}$, $\Delta$,
total number of mini-batches ($T.R$), and sampling fraction ($B /
total\ dataset\ size$) -- remain the same throughout the training
process.  \localgroupdp\ enforces ($\mathcal{E}/Z,\Delta/(Z e^{(Z-1)
  \frac{\mathcal{E}}{Z}})$-differential privacy, which
by~\sref{Definition}{def:group-dp} implies
($\mathcal{E}$,$\Delta$)-group differential privacy, hence subject
level DP by~\sref{Theorem}{thm:group-sub-dp}.

\subsection{Hierarchical Gradient Averaging}
\label{sec:higradavg}

While \localgroupdp\ may seem like a reasonable approach to enforce
subject level DP, its utility penalty due to group DP can be
significant.  For instance, even a group of size $2$ effectively
\emph{halves} the available privacy budget $\mathcal{E}$ for training.
The key challenge to enforce subject level DP is that the following
constraint seems fundamental: \emph{To guarantee subject level DP, any
  training algorithm must obfuscate the entire contribution made by
  any subject in the model's parameter updates.}
\localgroupdp\ complies with this constraint by enforcing group DP.

Our new algorithm, called
\higradavg\ (\sref{Algorithm}{algo:hi-grad-avg-dp}), takes a
diametrically opposite view to comply with the same constraint:
Instead of scaling the noise to a subject's group size (as is done in
\localgroupdp), \higradavg\ \emph{scales down} each subject's
mini-batch gradient contribution to the clipping threshold $C$.  This
is done in three steps: (i) collect data items belonging to a common
subject in the sampled mini-batch, (ii) compute and clip gradients
using the threshold $C$ for each individual data item of the subject,
and (iii) average those clipped gradients for the subject, denoted by
$g_a$.  Clipping and then averaging gradients ensures that the entire
subject's gradient norm is bounded by $C$.

Subsequently, \higradavg\ sums all the per-subject averaged gradients
along with the Gaussian noise, which are then averaged over the number
of distinct subjects $|subjects(S)|$ sampled in the mini-batch $S$.
\higradavg\ gets its name from this average-of-averages step.  These
two averaging steps have the result of mapping each subject $a$'s data
items' gradients to a single representative averaged gradient for $a$
in the mini-batch $S$.

\begin{algorithm}[t]

  {
    
    \KwInput{Set of $n$ users $\mathcal{U} = {u_i,u_2,...,u_n}$;
      $\mathcal{D}_i$, the dataset of user $u_i$; $M$, the model to be
      trained; $\theta$, the parameters of model $M$ ; gradient norm
      bound $C$; noise scale $\sigma$; sample of users $U_s$;
      mini-batch size $B$; $R$ training rounds; $T$ batches per round;
      $\eta$ the learning rate; $\mathcal{S}^{\mathcal{B}}_a$ the
      subset of data items from set $\mathcal{B}$ that have $a$ as
      their subject.
    }

  \setlength{\columnsep}{1pt}

  \SetKwProg{Fn}{}{}{}
  \Fn{HiGradAvgDP($u_i$):}{
    \For{$t = 1$ \KwTo $T$}{
      $\mathcal{B}$ = random sample of $B$ data items from $\mathcal{D}_i$\\
      \For{$a \in subjects(\mathcal{B})$}{
        \For{$s_i \in \mathcal{S}^{\mathcal{B}}_a$}{
          {\bf Compute gradients:}\\
          $g(s_i) = \triangledown \mathcal{L}(\theta,s_i)$\\
          {\bf Clip gradients:}\\
          $\bar{g}(s_i) = Clip(g(s_i),C)$\\
        }
            {\bf Average subject $a$'s gradients:}\\
            $g_a = \frac{1}{|\mathcal{S}^{\mathcal{B}}_a|} (\sum_i \bar{g}(s_i))$\\
      }
      $\tilde{g}_{\mathcal{B}} = \frac{ (\sum_{a \in subjects(\mathcal{B})} g_a + \mathcal{N}(0, \sigma^2 C^2 {\bf I}))}{|subjects(\mathcal{B})|}$\\
      $\theta = \theta - \eta \tilde{g}_{\mathcal{B}}$\\
    }
        {\bf return} $M$\\
  }
  \SetKwProg{Fn}{}{}{}
  \Fn{Server Loop:}{
    \For{$r = 1$ \KwTo $R$}{
      $U_s$ = sample $s$ users from $\mathcal{U}$\\
      \For{$u_i \in U_s$}{
        $\theta_i = \emph{HiGradAvgDP}(u_i$)\\
      }
      $\theta = \frac{1}{s} \sum_i \theta_i$ \\
      Send $M$ to all users in $\mathcal{U}$\\
    }
  }
  }
  \caption{Pseudo code for \higradavg\ that guarantees \emph{subject
      level} DP via hierarchical gradient averaging.}
\label{algo:hi-grad-avg-dp}
\end{algorithm}

The Gaussian noise scale $\sigma$ is calculated independently at each
user $u_i$ using standard parameters -- the privacy budget
$\varepsilon$, the failure probability $\delta$ and total number of
mini-batches $T.R$ over the entire multi-round training process.  For
the sampling fraction, we must consider sampling probability of
individual \emph{subjects} instead of data items.  As a result, the
subject sampling fraction becomes $q = \frac{k B}{|D_i|}$, where $k$
is the maximum number of data items belonging to any subject $a \in
subjects(\mathcal{B})$ in the dataset $D_i$.  Composition of the
privacy loss is done using the moments accountant
method~\cite{abadi16}.

To formally prove that \higradavg\ enforces subject level DP, we first
provide a formal definition of \emph{subject sensitivity} in a sampled
mini-batch.

\begin{definition}[Subject Sensitivity]
  Given a model $\mathcal{M}$, and a sampled mini-batch $\mathcal{B}$
  of training data, we define \emph{subject sensitivity}
  $\mathbb{S}^{\mathcal{B}}$ of $\mathcal{B}$ as the upper bound on
  the gradient norm of any single subject $a \in
  subjects(\mathcal{B})$.
\label{def:subject-sensitivity}
\end{definition}

The per-subject averaging of clipped gradients $g_a$ results in the
following lemma
\begin{lemma}
  For every sampled mini-batch $\mathcal{B}$ in a sampled user $u_i$'s
  training round in \higradavg, the subject sensitivity
  $\mathbb{S}^{\mathcal{B}}$ for $\mathcal{B}$ is bounded by $C$;
  i.e. $\mathbb{S}^{\mathcal{B}} \leq \vert C \vert$.
\label{lemma-hi-grad-avg-dp}
\end{lemma}

Scaling the Gaussian noise parameter $\sigma$ by a factor of $|C|$
ensures that the noise matches any subject's signal in each
mini-batch.  Furthermore, $\sigma$ itself is derived, based
on~\autoref{thm:dpsgd}, from
\begin{theorem}
  There exist constants $c_1$ and $c_2$ such that given the
  \emph{subject sampling} probability $q = \frac{kB}{|D_i|}$, where
  $k$ is the expected number of data items per subject in dataset
  $D_i$, and number of steps $T$, for any $\varepsilon < c_1 k^2 q^2
  T$, \higradavg\ enforces subject level
  ($\varepsilon$,$\delta$)-differential privacy for any $\delta > 0$
  if we choose
  \begin{align*}
    \sigma \ge c_2 \frac{k q \sqrt{T log(1/\delta)}}{\varepsilon}
  \end{align*}
\label{thm-hi-grad-avg-dp}
\end{theorem}
The proof for \sref{Theorem}{thm-hi-grad-avg-dp} follows the proof for
Theorem 1 in~\cite{abadi16} albeit by changing the sampling
probability from $q = \frac{B}{|D_i|}$ to $q = \frac{kB}{|D_i|}$.
The sampling probability's scaling factor $p$ captures sampling of data
subjects instead of data items.  As a result, the noise parameter
$\sigma$ scales up by a factor of $p$ for subject level DP as compared
to item level DP in~\cite{abadi16}.

\subsection{User Level Local Differential Privacy}
\label{sub:local-dp}

While centrally enforced user level privacy~\cite{mcmahan18} is not
sufficient to guarantee subject level privacy, we observe that
\emph{Local Differential Privacy
  (LDP)}~\cite{duchi13,kasiviswanathan08,warner65} is sufficient to
guarantee subject level privacy.  There are strong parallels between
the traditional LDP setting, where a data analyst can get access to
the data only after it has been perturbed, and privacy in the FL
setting, where the federation server gets access to parameter updates
from users after they have been locally perturbed by the users.  In
fact, LDP obfuscates the entire signal from a user to the extent that
an adversary, even the federation server, cannot tell the difference
between the signals coming from any two different users.

\begin{definition}
We say that FL algorithm $\mathcal{A}:\mathcal{D_U} \rightarrow
\mathcal{R}$ is \emph{user level ($\varepsilon$,$\delta$)-locally
  differentially private}, where $\mathcal{D_U}$ is the dataset domain
of users in set $\mathcal{U}$, and $\mathcal{R}$ is the model
parameter domain, if for any two users $u_1,u_2 \in \mathcal{U}$, and
$S \subseteq \mathcal{R}$,
\begin{eqnarray}
  \mathrm{Pr}[\mathcal{A}(D_{u_1}) \in S] \leq e^\varepsilon \mathrm{Pr}[\mathcal{A}(D_{u_2}) \in S] + \delta
  \label{eq:user-level-local-dp}
\end{eqnarray}
where $D_{u_1}$ and $D_{u_2}$ are the datasets of users $u_1$ and
$u_2$ respectively.
\end{definition}

We now present a new user level ($\varepsilon$,$\delta$)-LDP algorithm
called \userlocaldpsgd.  The underlying intuition behind this
algorithm is to let the user locally inject enough noise to make its
entire signal \emph{indistinguishable} from any other user's signal.
In every training round, each federation user enforces user level LDP
independently of the federation and any other users in the federation.
The federation server simply averages parameter updates received from
users and broadcasts the new averaged parameters back to the users.

\setlength{\textfloatsep}{0pt}
\begin{algorithm}[t]
  {

  \KwInput{Set of $n$ users $\mathcal{U} = {u_i,u_2,...,u_n}$;
    $\mathcal{D}_i$, the dataset of user $u_i$; $M$, the model to be
    trained; $\theta$, the parameters of model $M$ ; noise scale
    $\sigma$; gradient norm bound $C$; mini-batch size $B$; $R$
    training rounds; the learning rate $\eta$.
  }


    \setlength{\columnsep}{1pt}
    \emph{UserLDP($u_i$):\\}
    \SetKwProg{Fn}{}{}{}
      \For{$t = 1$ \KwTo $T$}{
        $\mathcal{B}$ = random sample of $B$ data items from $\mathcal{D}_i$\\
        {\bf Compute gradients:}\\
        $g(\mathcal{B}) = \triangledown \mathcal{L}(\theta,\mathcal{B})$\\
        {\bf Clip gradients:}\\
        $\bar{g}(\mathcal{B}) = g(\mathcal{B}) / max(1,\frac{\Vert g(\mathcal{B}) \Vert_2}{C})$\\
        {\bf Add Gaussian noise:}\\
        $\tilde{g}(\mathcal{B}) =  \bar{g}(\mathcal{B}) + \mathcal{N}(0,\sigma^2 C^2 {\bf I})$\\
        $\theta = \theta - \eta \tilde{g}(\mathcal{B})$\\
      }
      {\bf return} $\theta$
    \setlength\columnsep{1pt}

    \emph{Server Loop:}\\
    \SetKwProg{Fn}{}{}{}
      \For{$r = 1$ \KwTo $R$}{
        $U_s$ = sample $s$ users from $\mathcal{U}$\\
        \For{$u_i \in U_s$}{
          $\theta_i = \emph{UserDPSGD}(u_i)$\\
        }
        $\theta = \frac{1}{s} \sum_{i=1}^s \theta_i$\\
        Send $M$ to all users \\\ \ \ \ \ \ \ \ \ \ \ in $\mathcal{U}$\\
      }
    
  }
  \caption{Pseudo code for \userlocaldpsgd.
  }
\label{algo-user-local-dp-sgd}
\end{algorithm}

\userlocaldpsgd{}'s pseudo code appears
in~\sref{Algorithm}{algo-user-local-dp-sgd}.  Note that
\userlocaldpsgd\ appears very similar to \dpsgd~\cite{abadi16}.
However, there are two significant differences between the two
algorithms: First, while \dpsgd\ scales the noise proportional to the
gradient contribution of any single data item in a mini-batch,
\userlocaldpsgd\ computes noise proportional to the gradient
contribution of the entire mini-batch (line $9$).  To guarantee DP, we
need to first cap the sensitivity of each user $u_i$'s contribution to
parameter updates.  To that end, we focus on change affected by any
mini-batch $\mathcal{B}$ trained at $u_i$.  Line $7$
in~\sref{Algorithm}{algo-user-local-dp-sgd} ensures the following lemma.

\begin{lemma}
  For every mini-batch $\mathcal{B}$ of a sampled user $u_i$'s
  training round in \userlocaldpsgd, the sensitivity
  $\mathbb{S}_{\mathcal{B}}$ of the computed parameter gradient is
  bounded by $C$; i.e. $\mathbb{S}_{\mathcal{B}} \leq \vert C \vert$.
\label{lemma-user-local-dp-sgd}
\end{lemma}

Second, since we are interested in enforcing user level LDP, the
sampling probability of the user $u_i$ for each of its mini-batches is
$q = 1$.  Thus the Gaussian noise parameter $\sigma$ is derived, again
based on~\autoref{thm:dpsgd}, from
\begin{theorem}
  There exist constants $c_1$ and $c_2$ such that given the number of
  steps $T$, for any $\varepsilon < c_1 T$, \userlocaldpsgd\ enforces
  user level ($\varepsilon$,$\delta$)-differential privacy for any
  $\delta > 0$ if we choose
  \begin{align*}
    \sigma \ge c_2 \frac{\sqrt{T log(1/\delta)}}{\varepsilon}
  \end{align*}
\label{thm:sigma-for-userldp}
\end{theorem}

Sampling probability of $q = 1$ precludes privacy amplification by
sampling~\cite{abadi16,bassily14,kasiviswanathan08,wang19b}, which
significantly degrades the trained model's utility.

For any sampled user $u_i$, assume w.l.o.g. that $u_i$ trains for $T$
mini-batches in a single training round.  Thus the aggregate
sensitivity of parameter updates over a training round ($T$
mini-batches) for $u_i$ is bounded by $\eta T C$, where $\eta$ is the
mini-batch learning rate.  Thus the parameter update from $u_i$, as
observed by the federation server, is norm bounded by $\eta T C$, and
the cumulative noise from the distribution $\mathcal{N}(0,\eta T \sigma^2
C^2 {\bf I}$) (by linear composition of Gaussian distributions).  More
precisely, let $\mathcal{C}$ be the change in parameters affected by
any user $u_i$.  Then
\begin{equation}
  \vert\vert \mathcal{C}(u_i) \vert\vert_2 \leq
  \eta (\vert TC \vert + \vert \mathcal{N}(0,T \sigma^2 C^2 {\bf I} \vert)
\label{eq:randomized-response}
\end{equation}

\sref{Lemma}{lemma-user-local-dp-sgd}
and~\autoref{thm:sigma-for-userldp} ensure that the parameter update
signal for the entire training round at $u_i$ is matched with
correctly calibrated Gaussian noise forming a locally \emph{randomized
  response}~\cite{kasiviswanathan08,warner65} that is shared with the
federation server.

\begin{theorem}
  \userlocaldpsgd\ with parameter updates
  satisfying~\autoref{eq:randomized-response} as observed by the
  federation server in a training round, enforces user level
  ($\varepsilon$,$\delta$)-local differential privacy provided the
  noise parameter $\sigma$ satisfies the inequality
  from~\autoref{thm:sigma-for-userldp}, and the following inequality
  \begin{align*}
    \sigma > \frac{1}{\sqrt{2 \pi} \varepsilon \delta e^{\varepsilon}}
  \end{align*}
\label{thm-user-local-dp-sgd}
\end{theorem}

Proof for~\autoref{thm-user-local-dp-sgd} appears in the appendix.

\subsection{Composition Over Multiple Training Rounds}

Composition of privacy loss across multiple training rounds can be
done by straightforward application of DP composition results, such as
the moments accountant method that we use in our work.  Thus the
privacy loss $\varepsilon_r$ incurred in any single training round $r$
amplifies by a factor of $\sqrt{R}$ when federated training runs for
$R$ rounds.  We note that privacy losses are incurred by federation
users independently of other federation users.  Foreknowledge of the
number of training rounds $R$ lets us calculate the Gaussian noise
distribution's standard deviation $\sigma$ for a privacy loss budget
of ($\varepsilon$,$\delta$) for the aggregate training over $R$
rounds.  Given an aggregate privacy loss budget of $\varepsilon$,
since all users train for an identical number of rounds $R$, they
incur a privacy loss of $\varepsilon_r = \frac{\varepsilon}{\sqrt{R}}$
in each training round $r$.  Notably, this privacy loss per training
round is the the same for all users even if their dataset
cardinalities are dramatically different.

\subsection{Composing Subject Level DP Across Federation Users}

At the beginning of a training round $r$, each sampled user receives a
copy of the global model, with parameters $\Theta_{r-1}$, which it
then retrains using its private data.  Since all sampled users start
retraining from the same model $\mathcal{M}_{\Theta_{r-1}}$, and
independently retrain the model using their respective private data,
parallel composition of privacy loss across these sampled users may
seem to apply naturally~\cite{mcsherry09}.  In that case, the
aggregate privacy loss incurred across multiple federation users, via
an aggregation such as federated averaging, remains identical to the
privacy loss $\varepsilon_r$ incurred individually at each user.
However, parallel composition was proposed for item level privacy,
where an item belongs to at most one participant.  With subject level
privacy, a subject's data items can span across multiple users, which
limits application of parallel privacy loss composition to only those
federations where each subject's data is restricted to at most one
federation user.  In the more general case, we show that subject level
privacy loss composes \emph{adaptively} via the federated averaging
aggregation algorithm used in our FL training algorithms.

Formally, consider a FL training algorithm
$\mathcal{F}=(\mathcal{F}_l,\mathcal{F}_g)$, where $\mathcal{F}_l$ is
the user local component, and $\mathcal{F}_g$ the global aggregation
component of $\mathcal{F}$.  Given a federation user $u_i$, let
$\mathcal{F}_l: (\mathcal{M},D_{u_i}) \rightarrow \theta_{u_i}$, where
$\mathcal{M}$ is a model, $D_{u_i}$ is the private dataset of user
$u_i$, and $\theta_{u_i}$ is the updated parameters produced by
$\mathcal{F}_l$.  Let $\mathcal{F}_g$ be $\frac{1}{n} \sum_i
\theta_{u_i}$, a parameter update averaging algorithm over a set of
$n$ federation users $u_i$.

\begin{theorem}
  Given a FL training algorithm
  $\mathcal{F}=(\mathcal{F}_l,\mathcal{F}_g)$, in the most general
  case where a subject's data resides in the private datasets of
  multiple federation users $u_i$, the aggregation algorithm
  $\mathcal{F}_g$ adaptively composes subject level privacy losses
  incurred by $\mathcal{F}_l$ at each federation user.
  \label{thm-horizontal-composition}
\end{theorem}

We term this composition of privacy loss across federation users as
\emph{horizontal composition}.  Horizontal composition has a
significant effect on the number of federated training rounds
permitted under a given privacy loss budget.

\begin{theorem}
  Consider a FL training algorithm
  $\mathcal{F}=(\mathcal{F}_l,\mathcal{F}_g)$ that samples $s$ users
  per training round, and trains the model $\mathcal{M}$ for $R$
  rounds.  Let $\mathcal{F}_l$ at each participating user, over the
  aggregate of $R$ training rounds, locally enforce subject level
  ($\varepsilon$,$\delta$)-DP.  Then $\mathcal{F}$ globally enforces
  the same subject level ($\varepsilon$,$\delta$)-DP guarantee by
  training for $\frac{R}{\sqrt{s}}$ rounds.
  \label{thm-horizontal-composition-rounds}
\end{theorem}

The main intuition behind~\autoref{thm-horizontal-composition-rounds}
is that the $s$-way horizontal composition via $\mathcal{F}_g$ results
in an increase in training mini-batches by a factor of $s$.  As a
result, the privacy loss calculated by the moments accountant method
amplifies by a factor of $\sqrt{s}$, thereby forcing a reduction in
number of training rounds by a factor of $\sqrt{s}$ to counteract the
privacy loss amplification.  This reduction in training rounds can
have a significant impact on the resulting model's performance, as we
demonstrate in~\autoref{sec:evaluation}.
Proofs for~\sref{Theorem}{thm-horizontal-composition}
and~\sref{Theorem}{thm-horizontal-composition-rounds} appear in the
appendix.

An alternate approach to account for horizontal composition of privacy
loss is to simply scale the number of training minibatches (called
\emph{lots} by Abadi et al.~\cite{abadi16}) by the number of
federation users sampled in each training round.  The scaled minibatch
(lot) count can be used by each user to privately calculate the noise
scale $\sigma$ at the beginning of the entire federated training
process.  An increase in the number of total minibatches does lead to
a significant increase in the noise introduced in each minibatch's
gradients, resulting in model performance degradation.

\section{Utility Loss}
\label{sec:utility-loss}

Our utility loss formalism leverages a long line of former work on
differentially private empirical risk minimization
(ERM)~\cite{bassily19,bassily14,chaudhuri11,duchi13,iyengar19,kifer12,song13,talwar15,thakurta13,wang18}.
In particular, we extend the notation of, and heavily base our formal
analysis on work by Bassily et al.~\cite{bassily19}, applying it to
subject level DP in general, with specializations for our individual
algorithms.

Let $\mathcal{Z}$ denote the data domain, and $\mathcal{D}$ denote a
data distribution over $\mathcal{Z}$.  We assume a $L$-Lipschitz
convex loss function $\ell: \mathbb{R}^d \times \mathcal{Z}
\rightarrow \mathbb{R}$ that maps a parameter vector ${\bf w} \in W$,
where $W \subset \mathbb{R}^d$ is a convex parameter space, and a data
point $z \in \mathcal{Z}$, to a real value.

\begin{definition}[$\alpha$-Uniform Stability~\cite{bassily19,bousquet02}]
  Let $\alpha > 0$.  A randomized algorithm $\mathcal{A}:
  \mathcal{Z}^n \rightarrow W$ is $\alpha$-uniformly stable (w.r.t. loss
  $\ell: W \times \mathcal{Z} \rightarrow \mathbb{R}$) if for any pair
  $S, S' \in \mathcal{Z}^n$ differing in at most one data point, we
  have
  \begin{align*}
    \sup_{z \in \mathcal{Z}} \underset{\mathcal{A}}{\mathbb{E}}
        [\ell(\mathcal{A}(S),z) - \ell(\mathcal{A}(S'),z)] \le \alpha
  \end{align*}  
  \label{def:uniform-stability}
\end{definition}

\begin{definition}[$(k,\alpha)$-Uniform Stability]
  Let $\alpha > 0$.  A randomized algorithm $\mathcal{A}:
  \mathcal{Z}^n \rightarrow W$ is said to be $(k,\alpha)$-uniformly
  stable (w.r.t loss $\ell: W \times \mathcal{Z} \rightarrow
  \mathbb{R}$) if for any pair $S, S' \in \mathcal{Z}^n$ differing in
  at most $k$ data points, we have
  \begin{align*}
    \sup_{z \in \mathcal{Z}} \underset{\mathcal{A}}{\mathbb{E}}
        [\ell(\mathcal{A}(S),z) - \ell(\mathcal{A}(S'),z)] \le k \alpha
  \end{align*}  
  \label{def:k-uniform-stability}  
\end{definition}

We use $(k,\alpha)$-uniform stability to represent the effect of a
data subject with cardinality $k$ in the dataset.  Thus algorithm
$\mathcal{A}$ is $(k,\frac{\beta}{k})$-uniformly stable if

\begin{align*}
  \underset{\mathcal{A}}{\mathbb{E}}[\ell(\mathcal{A}(S_k),z) -
    \ell(\mathcal{A}(S_0), z)] \le \beta
\end{align*}

\begin{lemma}
 A (randomized) algorithm $\mathcal{A}: \mathcal{Z}^n \rightarrow W$
 is $(k,\frac{\beta}{k})$-uniformly stable \emph{iff} it is
 $\frac{\beta}{k}$-uniformly stable.
\end{lemma}

\begin{proof}
  Consider sets $S_0,S_1,S_2,...,S_k \subset \mathcal{Z}$ such that
  $S_i = S_{i-1} \cup \{x_i\}$, for all $1 \le i \le k$, where $x_i
  \in \mathcal{Z}$.  In other words, $S_i$ contains a single
  additional data point than $S_{i-1}$.

  Assume that $\mathcal{A}: \mathcal{Z}^n \rightarrow W$ is
  $(k,\frac{\beta}{k})$-uniformly stable.  Then we have

  \begin{align*}
    \underset{\mathcal{A}}{\mathbb{E}} [\ell(\mathcal{A}(S_k),z) -
      \ell(\mathcal{A}(S_0),z)]\\
    &\hspace{-1in}= \underset{\mathcal{A}}{\mathbb{E}} [\sum_{i=1}^{k}
      (\ell(\mathcal{A}(S_i),z) - \ell(\mathcal{A}(S_{i-1}),z))]\\
    &\hspace{-1in}= \sum_{i=1}^{k} \underset{\mathcal{A}}{\mathbb{E}}
           [\ell(\mathcal{A}(S_i),z) - \ell(\mathcal{A}(S_{i-1}),z)]\\
    &\hspace{-1in}\le \beta
  \end{align*}

  By i.i.d. and symmetry assumptions, we get $\forall i \in
  {1,2,...,k}$
  \begin{align*}
    \underset{\mathcal{A}}{\mathbb{E}} [\ell(\mathcal{A}(S_i),z) -
      \ell(\mathcal{A}(S_{i-1}),z)] \le \frac{\beta}{k}
  \end{align*}

  For the other direction of the \emph{iff} we use the same sets
  $S_0,S_1,S_2,...,S_k$, and assume $\forall i \in {1,2,...,k}$
  \begin{align*}
    \underset{\mathcal{A}}{\mathbb{E}} [\ell(\mathcal{A}(S_i),z) -
      \ell(\mathcal{A}(S_{i-1}),z)] \le \frac{\beta}{k}
  \end{align*}

  Hence,
  \begin{align*}
    \underset{\mathcal{A}}{\mathbb{E}} [\ell(\mathcal{A}(S_k),z) -
      \ell(\mathcal{A}(S_0),z)]\\    
    &\hspace{-1in}=\underset{\mathcal{A}}{\mathbb{E}} [\sum_{i=1}^{k}
      (\ell(\mathcal{A}(S_i),z) - \ell(\mathcal{A}(S_{i-1}),z))]\\
    &\hspace{-1in}= \sum_{i=1}^{k} \underset{\mathcal{A}}{\mathbb{E}}
    [\ell(\mathcal{A}(S_i),z) - \ell(\mathcal{A}(S_{i-1}),z)]\\
    &\hspace{-1in} \le \beta
  \end{align*}
  
\end{proof}

\subsection{General Utility Loss for Subject Level Privacy}

Given the parameter vector ${\bf w} \in W$, dataset $S =
{s_1,s_2,...,s_n}$, and loss function $\ell$, we define the
\emph{empirical loss} of ${\bf w}$ as $\hat{\mathcal{L}}({\bf w}; S)
\triangleq \frac{1}{n} \sum_{i=1}^n \ell({\bf w},s_i)$, and the
\emph{excess empirical loss} of ${\bf w}$ as $\Delta
\hat{\mathcal{L}}({\bf w};S) \triangleq \hat{\mathcal{L}}({\bf w};S) -
\underset{\tilde{\bf w} \in W}{min} \hat{\mathcal{L}}(\tilde{\bf
  w};S)$.  Similarly we define the \emph{population loss} of ${\bf w}
\in W$ w.r.t. loss $\ell$ and a distribution $\mathcal{D}$ over
$\mathcal{Z}$ as $\mathcal{L}({\bf w};\mathcal{D}) \triangleq
\underset{z \sim \mathcal{D}}{\mathbb{E}} [\ell({\bf w},z)]$.  The
excess population loss of ${\bf w}$ is defined as $\Delta
\mathcal{L}({\bf w};\mathcal{D}) \triangleq \mathcal{L}({\bf
  w};\mathcal{D}) - \underset{\tilde{\bf w} \in W}{min}
\mathcal{L}(\tilde{\bf w}; \mathcal{D})$.

\begin{lemma}[from~\cite{bassily19}]
  Let $\mathcal{A}: \mathcal{Z}^n \rightarrow W$ be a
  $\frac{\beta}{k}$-uniformly stable algorithm w.r.t. loss $\ell: W \times
  \mathcal{Z} \rightarrow \mathbb{R}$.  Let $\mathcal{D}$ be any
  distribution over $\mathcal{Z}$, and let $S \sim \mathcal{D}^n$.  Then,
  \begin{equation}
    \underset{S \sim \mathcal{D}^n,
      \mathcal{A}}{\mathbb{E}}[\mathcal{L}(\mathcal{A}(S);
      \mathcal{D}) - \hat{\mathcal{L}}(\mathcal{A}(S); S)] \le \frac{\beta}{k}
    \label{eq:uniform-stability-bound}
  \end{equation}
  \label{lemma:uniform-stability-bound}
\end{lemma}

Let $\mathcal{A}$ be a $L$-Lipschitz convex function that uses dataset
$S$ to generate an approximate minimizer $\hat{\bf w}_S \in W$ for
$\mathcal{L}(.;\mathcal{D})$.  Thus the accuracy of $\mathcal{A}$ is
measured in terms of \emph{expected} excess population loss

\begin{equation}
  \Delta \mathcal{L}(\mathcal{A}; \mathcal{D}) \triangleq \mathbb{E} [
    \mathcal{L}(\hat{\bf w}_S; \mathcal{D}) - \underset{{\bf w} \in
      W}{min} \mathcal{L}({\bf w}; \mathcal{D})]
  \label{eq:expected-excess-population-loss}
\end{equation}

\begin{lemma}
  Let $\mathcal{A}_{SDP}$ be a $L$-Lipschitz randomized algorithm
  that guarantees subject level $(\varepsilon,\delta)$-DP.
  Let $T$ be the number of training iterations, $m$ the minibatch size
  per training step, and $\eta$ the learning rate.  Then,
  $\mathcal{A}_{SDP}$ is $(\kappa,\alpha)$-uniformly stable, where
  $\kappa$ is the expected number of data items for any subject $s_p$
  appearing in $\mathcal{A}_{SDP}$'s training dataset, and $\alpha =
  L^2 \frac{(T + 1) \eta}{n}$.
  \label{lemma:a-sdp-uniform-stability}
\end{lemma}

\begin{proof}
  Consider dataset $S$ comprising data items of $n_s$ subjects
  $s_1,...,s_{p-1},s_p,s_{p+1},...,s_{n_s}$, and dataset $S'$
  comprising data items of $n_s-1$ subjects
  $s_1,...,s_{p-1},s_{p+1},...,s_{n_s}$; i.e. $S$ and $S'$ differ from
  each other by a single data subject $s_p$.  Let number of data items
  per subject $|s_i| > 0$.

  Let ${\bf w}_0,{\bf w}_1,...{\bf w}_T$ and ${\bf w}'_0,{\bf
    w}'_1,...,{\bf w}'_T$ be the parameter values of
  $\mathcal{A}_{SDP}$ corresponding to $T$ training steps taken over
  input datasets $S$ and $S'$ respectively.  Let $\xi_t \triangleq
  {\bf w}_t - {\bf w}'_t$ for any $t \in [T]$.
  
  Assume random sampling with replacement for a minibatch of data
  items.  Let $r$ be the number of data items in a sampled minibatch
  of size $m$ that belong to subject $s_p$.  Then, by the
  non-expansiveness property of the gradient update step, we have

  \begin{align*}
    \|\xi_{\tau + 1}\| \le \|\xi_{\tau}\| + 2 L \eta \frac{r}{m}
  \end{align*}

  Note that $r$ is a binomial random variable.  Thus the expected
  value of $r$, i.e. $\mathbb{E}[r] = m \frac{\kappa}{n}$, where
  $\kappa = \mathbb{E}[|s_p|]$.  Thus $\kappa$ depends on the
  underlying data distribution $\mathcal{D}$.  For instance, if
  $\mathcal{D}$ is a uniform distribution, $\kappa = \frac{n}{n_s}$,
  where $n_s$ is the number subjects in $S$.  Assuming $\| \xi_0 \| =
  0$, taking expectation and using the induction hypothesis, we get

  \begin{align*}
    \mathbb{E}[\|\xi_{\tau + 1}\|]
    &\le 2 L \frac{\eta (\tau +
      1) \kappa}{n}\\
    &= 2 L \frac{\eta (\tau + 1) \kappa}{n}
  \end{align*}

  Now let $\bar{\bf w}_T = \sum_{t=1}^T {\bf w}_t$ and $\bar{\bf w}'_T
  = \sum_{t=1}^T {\bf w}'_t$.  Since $\ell$ is $L$-Lipschitz, for
  every $z \in \mathcal{Z}$ we get

  \begin{align*}
    \mathbb{E}[\ell(\bar{\bf w}_T, z) - \ell(\bar{\bf w}'_T, z)]
    &= \mathbb{E}[\ell(\mathcal{A}_{SDP}(S), z) - \ell(\mathcal{A}_{SDP}(S'), z)] \\
    & \le L.\mathbb{E}[\|\bar{\bf w}_T - \bar{\bf w}'_T\|] \\
    & \le L \frac{1}{T} \sum_{t=1}^T \mathbb{E}[\|\xi_t\|] \\
    & \le L \frac{1}{T} \frac{2 L \eta \kappa}{n} \sum_{t=1}^T t \\
    & = L^2 \frac{(T+1) \eta \kappa}{n} 
  \end{align*}
  
\end{proof}

Note that the above bound is a scaled version (by $\kappa$) of the
recently shown bound for item level DP~\cite{bassily19}.  Thus,
intuitively in our case, the smaller the number of data items per
subject in a dataset, the closer our bound is to that of item level
DP.  Our bound is identical to the item level DP bound in the extreme
case where each subject has just one data item in the dataset.

From~\sref{Lemma}{lemma:a-sdp-uniform-stability},
~\autoref{eq:uniform-stability-bound}
and~\autoref{eq:expected-excess-population-loss}, and substituting $k
= \kappa$ and $\beta = L^2 \frac{(T+1) \eta \kappa}{n}$
in~\autoref{eq:uniform-stability-bound}, we get

\begin{theorem}
  Let $\mathcal{A}_{SDP}$ be a $L$-Lipschitz randomized algorithm that
  guarantees subject level $(\varepsilon,\delta)$-DP.  Then its excess
  population loss is bounded by

  \begin{align*}
    \Delta \mathcal{L}(\mathcal{A}_{SDP}; \mathcal{D}) \le \\
    & \hspace{-0.75in} \underset{S \sim \mathcal{D}^n,
      \mathcal{A}_{SDP}}{\mathbb{E}}[\hat{\mathcal{L}}(\bar{\bf
        w}_T;S) - \underset{{\bf w} \in W}{min}\hat{\mathcal{L}}({\bf
        w}; S)] + L^2 \frac{\eta (T+1)}{n}
  \end{align*}
  \label{thm:excess-population-loss}
\end{theorem}

Interestingly, the above inequality appears to be identical to the
excess population loss bound of work by Bassily et al. on item level
DP~\cite{bassily19}.  However, only the third RHS term is identical,
and the first two RHS terms evaluate to different quantities for all
of our algorithms as we show below.

\subsection{Utility Loss for \localgroupdp\ and \userlocaldpsgd}

We now formally show how \localgroupdp\ amplifies the Gaussian noise
that factors directly into excess population loss $\Delta
\mathcal{L}$.

\begin{lemma}
  Let $W$ be the $M$-bounded convex parameter space for \localgroupdp,
  and $S \in \mathcal{Z}^n$ be the input (training) dataset.  Let
  $(\varepsilon,\delta)$ be the subject level DP parameters for
  \localgroupdp, $q$ be the minibatch sampling ratio, and $d$ the
  model dimensionality.  Then, for any $\eta > 0$, the excess
  empirical loss of \localgroupdp\ is bounded by

  \begin{align*}
    \mathbb{E}[\hat{\mathcal{L}}(\bar{\bf w}_T;S)] - \underset{{\bf w}
      \in W}{min} \hat{\mathcal{L}}({\bf w};S) \le \\
    & \hspace{-1in} \frac{M^2}{2 \eta T} + \frac{\eta L^2}{2} + \eta d
    \frac{c_2^2 k^2 q^2}{\varepsilon^2} \left( T log \frac{k e^{(k-1)
        \varepsilon/k}}{\delta} \right)
  \end{align*}
  \label{lemma:localgroupdp-utility}
\end{lemma}

\begin{proof}
  From the classic analysis of gradient descent on convex-Lipschitz
  functions~\cite{bassily19,ssbd14}, we get

  \begin{align*}
    \mathbb{E}[\hat{\mathcal{L}}(\bar{\bf w}_T;S)] - \underset{{\bf w}
      \in W}{min} \hat{\mathcal{L}}({\bf w};S) \le \frac{M^2}{2 \eta
      T} + \frac{\eta L^2}{2} + \eta \sigma^2 d
  \end{align*}

  where the last term on the RHS of the inequality is the additional
  empirical error due to that privacy enforcing noise.

  By Theorem $1$ from~\cite{abadi16}, the term $\sigma$ is lower
  bounded by

  \begin{equation}
    \sigma \ge c_2 \frac{q \sqrt{T log(1/\delta)}}{\varepsilon}
    \label{eq:sigma}
  \end{equation}

  for item level DP.  Extending the bound to group level DP, for
  groups of size $k$, by substituting $(\varepsilon,\delta)$ with $(k
  \varepsilon,k e^{(k-1)\varepsilon} \delta)$ gives us

  \begin{align*}
    \sigma \ge c_2 \frac{k q \sqrt{T log(\frac{k e^{(k-1)
            \varepsilon/k}}{\delta})}}{\varepsilon}
  \end{align*}

  We get the theorem's inequality by substituting $\sigma$ as above.
  
\end{proof}

Combining~\sref{Lemma}{lemma:localgroupdp-utility}
with~\autoref{thm:excess-population-loss} gives us

\begin{theorem}
  The excess population loss of $\mathcal{A}_{\localgroupdp}$ is
  satisfied by

  \begin{align*}
    \Delta \mathcal{L}(\mathcal{A}_{\localgroupdp}; \mathcal{D}) \le \\
    & \hspace{-1in} \frac{M^2}{2 \eta T} + \frac{\eta L^2}{2} + \eta d
    \frac{c_2^2 k^2 q^2}{\varepsilon^2} \left( T log \frac{k e^{(k-1)
        \varepsilon/k}}{\delta} \right) \\
    & \hspace{-1in} + L^2 \frac{\eta (T + 1)}{n}
  \end{align*}
\end{theorem}

Note that the noise term amplifies quadratically with group size $k$,
which leads to rapid utility degradation with increasing group size.
The excess population loss measure for \userlocaldpsgd\ can be
obtained by simply replacing the group size term $k$ to the size of
the minibatch $m$, which clearly leads to significantly greater noise
amplification.  Furthermore, by~\autoref{thm:sigma-for-userldp}, the
noise parameter $\sigma$ is lower bounded by
\begin{align*}
  \sigma \ge c_2 \frac{\sqrt{T log(1/\delta)}}{\varepsilon}
\end{align*}

Recall the sampling probability escalates to $q = 1$ for
\userlocaldpsgd.  This leads to the following theorem

\begin{theorem}
  The excess population loss of $\mathcal{A}_{\userlocaldpsgd}$ is
  satisfied by

  \begin{align*}
    \Delta \mathcal{L}(\mathcal{A}_{\userlocaldpsgd}; \mathcal{D}) \le \\
    & \hspace{-1in} \frac{M^2}{2 \eta T} + \frac{\eta L^2}{2} + \eta d
    \frac{c_2^2 m^2}{\varepsilon^2} \left( T log \frac{m e^{(m-1)
        \varepsilon/m}}{\delta} \right) \\
    & \hspace{-1in} + L^2 \frac{\eta (T + 1)}{n}
  \end{align*}
\end{theorem}

\subsection{Utility Loss for \higradavg}

Recall that unlike \localgroupdp, \higradavg\ does not scale up the
noise to the group size of a subject in a minibatch.  It instead
scales down the gradients of all data items of the subject to a single
data item's gradient bounds (established by the clipping threshold).
As a result, the noise amplification we showed for \localgroupdp\ does
not exist for \higradavg.  However, scaling down the gradient signal
of a subject does indeed affect \higradavg{}'s utility.  To show the
effect formally we go back to the classic analysis of gradient descent
for convex-Lipschitz functions, Lemma $14.1$ in~\cite{ssbd14}.

Let ${\bf w}^* = \underset{{\bf w} \in
  W}{argmin}~\hat{\mathcal{L}}({\bf w},S)$.  Given that
$\hat{\mathcal{L}}$ is a convex $L$-Lipschitz function,
from~\cite{ssbd14} we have

\begin{equation}
  \mathbb{E}[\hat{\mathcal{L}}(\bar{\bf w}_T;S)] -
  \hat{\mathcal{L}}({\bf w}^*;S) \le \frac{1}{T} \sum_{t=1}^T
  \langle {\bf w}_t - {\bf w}^*, \bigtriangledown
  \hat{\mathcal{L}}({\bf w}_t)\rangle
  \label{eq:ssbd14}
\end{equation}

Consider ${\bf w}_{t+1} = {\bf w}_t + \eta {\bf v}_t$, where ${\bf
  v}_t = \bigtriangledown \hat{\mathcal{L}}({\bf w}_t)$, and $\eta$ is
the learning rate.

\begin{lemma}
  Consider algorithm $\mathcal{A}_{\higradavg}^-$ that performs the
  same steps as \higradavg\ except for the noise injection step (at
  line $12$ of Algorithm $3$).  Let
  $\hat{\mathcal{L}}_{\higradavg}^-({\bf w};S)$ be the $L$-Lipschitz
  continuous empirical loss function, and $W$ be the $M$-bounded
  convex parameter space for $\mathcal{A}_{\higradavg}^-$.  If $k$ is
  the expected number of data items per subject in a sampled
  minibatch, then

  \begin{align*}
    \mathbb{E}[\hat{\mathcal{L}}_{\higradavg}^-(\bar{\bf w}_T;S)] -
    \hat{\mathcal{L}}_{\higradavg}^-({\bf w}^*;S) \le
    \frac{k M^2}{2 \eta T} + \frac{\eta L^2}{2 k}
  \end{align*}
\end{lemma}

\begin{proof}
  Consider

  \begin{align*}
    \langle {\bf w}_t - {\bf w}^*, {\bf v}_t \rangle =
    \frac{k}{\eta} \langle {\bf w}_t - {\bf w}^*, \frac{\eta}{k}{\bf v}_t \rangle \\
    & \hspace{-2in}= \frac{k}{2 \eta} (- \|{\bf w}_t - {\bf w}^* - \frac{\eta}{k} {\bf v}_t \|^2 +
    \| {\bf w}_t - {\bf w}^* \|^2 + \frac{\eta^2}{k^2} \|{\bf v}_t \|^2) \\
    & \hspace{-2in}= \frac{k}{2 \eta} (- \|{\bf w}_{t+1} - {\bf w}^* \|^2 + \| {\bf w}_t - {\bf w}^* \|^2) +
    \frac{\eta}{2 k} \| {\bf v}_t \|^2
  \end{align*}

  Summing the equality over $t$ and collapsing the first term on the
  RHS gives us

  \begin{align*}
    \sum_{t=1}^T \langle {\bf w}_t - {\bf w}^*, {\bf v}_t \rangle \\
    & \hspace{-1in}= \frac{k}{2 \eta} (\| {\bf w}_1 - {\bf w}^* \|^2 - \| {\bf w}_{T+1} - {\bf w}^* \|^2) +
    \frac {\eta}{2 k} \sum_{t=1}^T \| {\bf v}_t \|^2 \\
    & \hspace{-1in} \le \frac{k}{2 \eta} (\| {\bf w}_1 - {\bf w}^* \|^2) +
    \frac{\eta}{2 k} \sum_{t=1}^T \| {\bf v}_t \|^2 \\
    & \hspace{-1in}= \frac{k}{2 \eta} (\| {\bf w}^* \|^2) +
    \frac{\eta}{2 k} \sum_{t=1}^T \| {\bf v}_t \|^2,
  \end{align*}
  assuming ${\bf w}_1 = 0$.  Since $W$ is $M$ bounded and
  $\hat{\mathcal{L}}_{\higradavg}^-$ is $L$-Lipschitz, combining the
  above with~\autoref{eq:ssbd14}, we get

  \begin{align*}
    \mathbb{E}[\hat{\mathcal{L}}_{\higradavg}^-(\bar{\bf w}_T;S)] -
    \hat{\mathcal{L}}_{\higradavg}^-({\bf w}^*;S) \le
    \frac{k M^2}{2 \eta T} + \frac{\eta L^2}{2 k}
  \end{align*}
\end{proof}

Now reintroducing the noise in \higradavg~(at line 12 in Algorithm
$3$), with $\hat{\mathcal{L}}_{\higradavg}({\bf w};S)$ as the
$L$-Lipschitz continuous loss function of \higradavg, we get
\begin{align*}
    \mathbb{E}[\hat{\mathcal{L}}_{\higradavg}(\bar{\bf w}_T;S)] -
    \hat{\mathcal{L}}_{\higradavg}({\bf w}^*;S) \le \\
    & \hspace{-1in}\frac{k M^2}{2 \eta T} + \frac{\eta L^2}{2 k} + \eta \sigma^2 d,
\end{align*}

where the last term of the RHS is the additional empirical
error due to the privacy enforcing noise~\cite{bassily19}.  Combining
the above inequality with~\autoref{thm-hi-grad-avg-dp}, we get
\begin{align*}
    \mathbb{E}[\hat{\mathcal{L}}_{\higradavg}(\bar{\bf w}_T;S)] -
    \hat{\mathcal{L}}_{\higradavg}({\bf w}^*;S) \le \\
    & \hspace{-1in}\frac{k^2 M^2 + \eta^2 T L^2}{2 k \eta T} +
    \eta d \frac{c_2^2 k^2 q^2}{\varepsilon^2} T log(1/\delta)
\end{align*}

Combining the above inequality
with~\autoref{thm:excess-population-loss} we get

\begin{theorem}
  The excess population loss of $\mathcal{A}_{\higradavg}$ is
  satisfied by

  \begin{align*}
    \Delta \mathcal{L}(\mathcal{A}_{\higradavg}; \mathcal{D}) \le
    & \frac{k^2 M^2 + \eta^2 T L^2}{2 k \eta T} + \eta d \frac{c_2^2 k^2
      q^2}{\varepsilon^2} T log(1/\delta) \\
    & + L^2 \frac{\eta (T+1)}{n}
  \end{align*}
\end{theorem}

The first term on the RHS of the inequality scales linearly with $k$,
the expected number of data items per subject.  Thus we should expect
some utility loss compared to \dpsgd\ for $k > 1$.  However, the
second noise term, scales quadratically with $k$, somewhat similar to
that in \localgroupdp\ and \userlocaldpsgd.

\section{Empirical Evaluation}
\label{sec:evaluation}

We implemented all our algorithms \userlocaldpsgd, \localgroupdp, and
\higradavg, and a version of the DP-SGD algorithm by Abadi et
al.~\cite{abadi16} that enforces item level DP in the FL setting
(\localdpsgd).  We also compare these algorithms with a FL training
algorithm, \fedavg~\cite{konecny15}, that does not enforce any privacy
guarantees.  All our algorithms are implemented in our distributed FL
framework built on distributed PyTorch.

We focus our evaluation on Cross-Silo FL~\cite{kairouz19}, which we
believe is the most appropriate setting for the subject level privacy
problem.  We use the FEMNIST and Shakespeare datasets~\cite{caldas18}
for our evaluation.  In FEMNIST, the hand-written numbers and letters
can be divided based on authors, which ordinarily serve as federation
users in FL experiments by most researchers.  In Shakespeare, each
character in the Shakespeare plays serves as a federation user.  In
our experiments however, the FEMNIST authors and Shakespeare play
characters are treated as data subjects.  To emulate the cross-silo FL
setting, we report evaluation on a 16-user federation.

We use the CNN model on FEMNIST appearing in the LEAF benchmark
suite~\cite{caldas18} as our target model to train.  More
specifically, the model consists of two convolution layers interleaved
with ReLU activations and maxpooling, followed by two fully connected
layers before a final log softmax layer.  For the Shakespeare dataset
we use a stacked LSTM model with two linear layers at the end.

We use $80\%$ of the training data for training, and $20\%$ for
validation.  Test data comes separately in FEMNIST and Shakespeare.
Training and testing was done on a local GPU cluster comprising $2$
nodes, each containing $8$ Nvidia Tesla V100 GPUs.

We extensively tuned the hyperparameters of mini-batch size $B$,
number of training rounds $T$, gradient clipping threshold $C$, and
learning rate $\eta$.  The final hyperparameters for FEMNIST were: $B
= 512$, $T = 100$, $C = 0.001$, and learning rates $\eta$ of $0.001$
and $0.01$ for the non-private and private FL algorithms respectively.
Shakespeare hyperparameters were: $B = 100$, $T = 200$, $C = 0.00001$,
and learning rates $\eta$ of $0.0002$ and $0.01$ for the non-private
and private FL algorithms.

In our implementations of all our algorithms \userlocaldpsgd,
\localgroupdp, and \higradavg, we used the privacy loss horizontal
composition accounting technique that reduces the number of training
rounds by $\sqrt{s}$, where $s$ is the number of sampled users per
training round.  We experimented with the alternative approach that
scales up the number of minibatches by $s$ to calculate a larger noise
scale $\sigma$, but this approach consistently yielded worse model
utility than our first approach.  Hence here we report only the
performance of our first approach.

\subsection{FEMNIST and Shakespeare Performance}

We first conduct an experiment that reports average test accuracy and
loss at the end of each training round, over a total of $100$ and
$200$ training rounds for FEMNIST and Shakespeare respectively.  The
FEMNIST dataset contains $3500$ subjects, and the Shakespeare dataset
contains $660$ subjects.  In FEMNIST, the average number of data items
per subject is $145$, whereas in case of Shakespeare it is $4,484$.
As we shall see later in this section, these subject cardinalities
significantly contribute to performance of our algorithms' models.
Each subject's data items are uniformly distributed among the $16$
federation users.

\begin{figure}
\centering
FEMNIST\\
\includegraphics[width=1.5in]{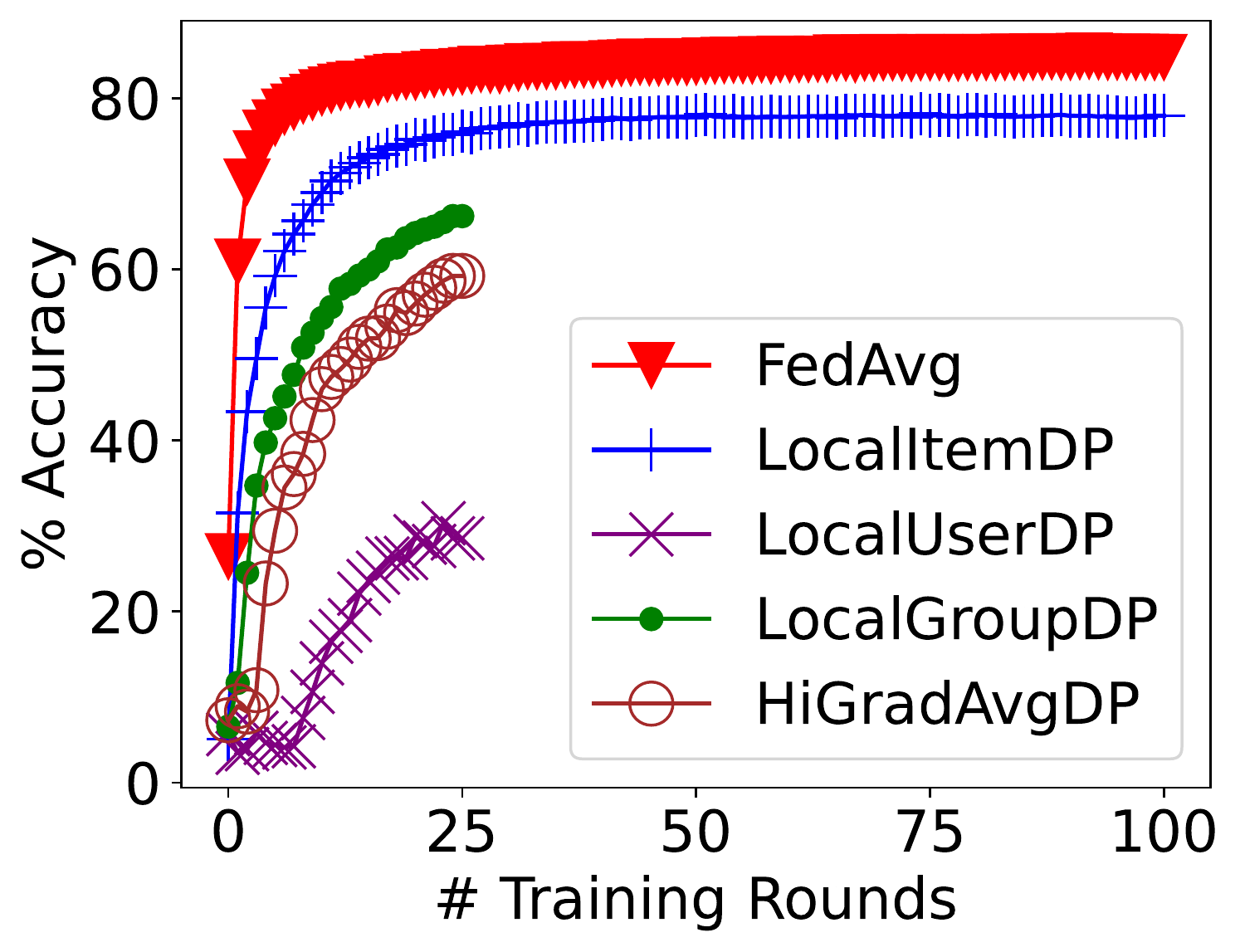}
\includegraphics[width=1.5in]{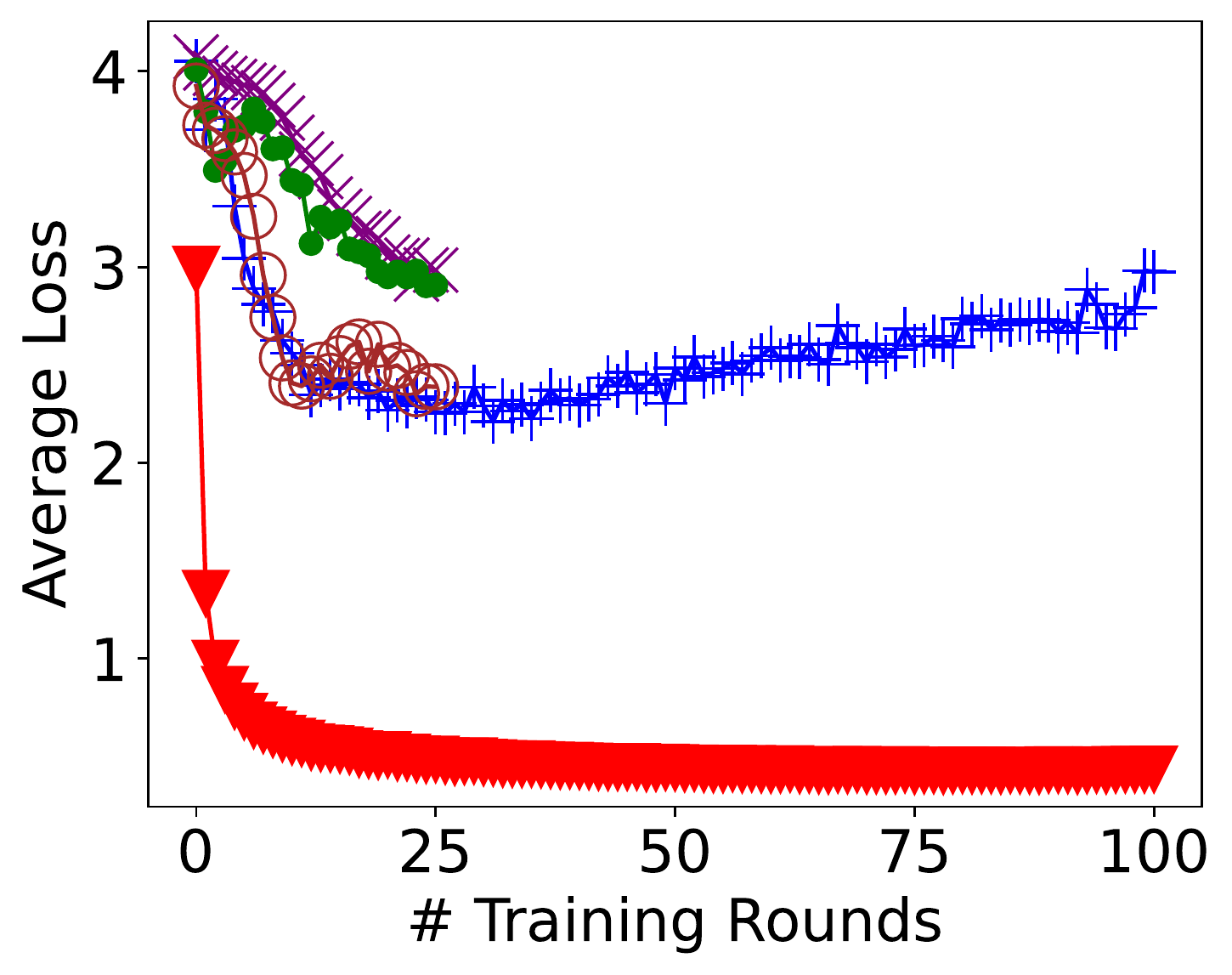}
\subfloat[{\hspace{0.4\linewidth}}\label{fig:femnist-accuracy}]{\hspace{.5\linewidth}}
\subfloat[\label{fig:femnist-loss}]{\hspace{.20\linewidth}}\\
Shakespeare\\
\includegraphics[width=1.5in]{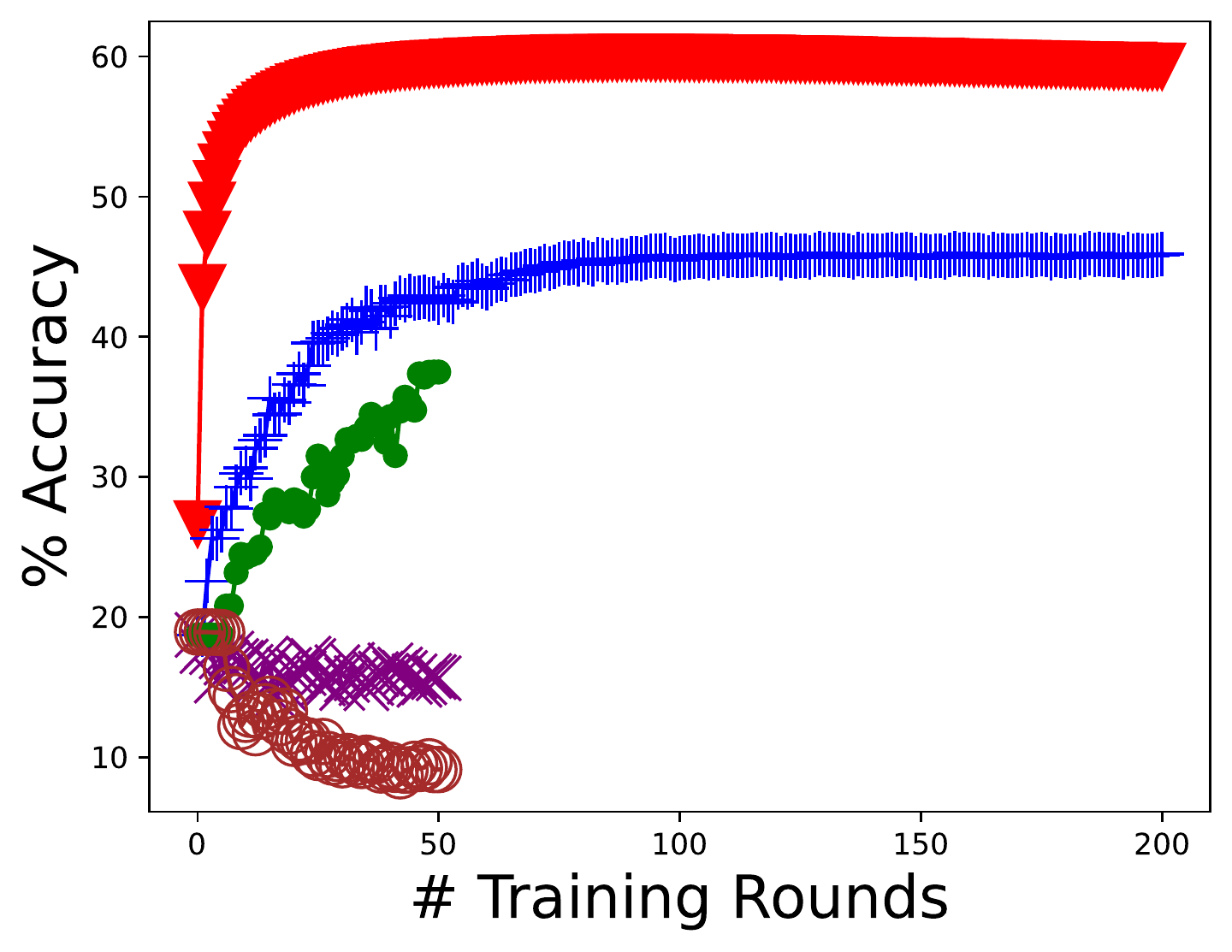}
\includegraphics[width=1.5in]{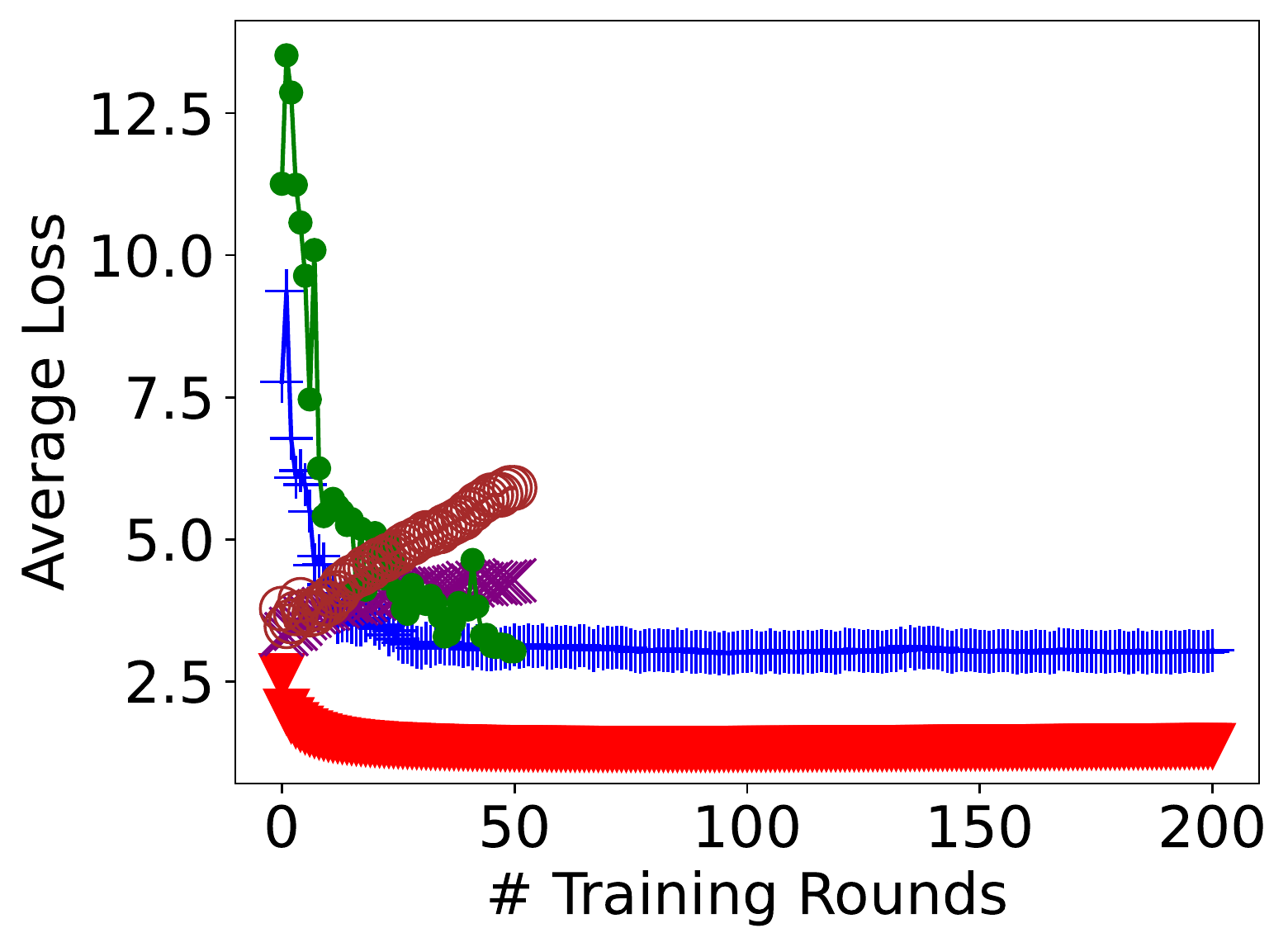}
\subfloat[{\hspace{0.4\linewidth}}\label{fig:shakespeare-accuracy}]{\hspace{.5\linewidth}}
\subfloat[\label{fig:shakespeare-loss}]{\hspace{.20\linewidth}}
\caption{\footnotesize Average test accuracy and loss on the FEMNIST
  (a),(b) and Shakespeare (c),(d) datasets over training rounds for
  various algorithms.  For DP guarantees: $\epsilon=4.0$ and
  $\delta=10^{-5}$ budgeted over all $100$ and $200$ training rounds
  for FEMNIST and Shakespeare respectively.  Model performance for the
  subject level privacy algorithms is constrained by the limited
  number of training rounds ($25$ for FEMNIST, and $50$ for
  Shakespeare) permitted under the prescribed privacy budget.
}
\label{fig:accuracy-loss}
\end{figure}

\begin{figure}
\centering
\includegraphics[width=2.0in]{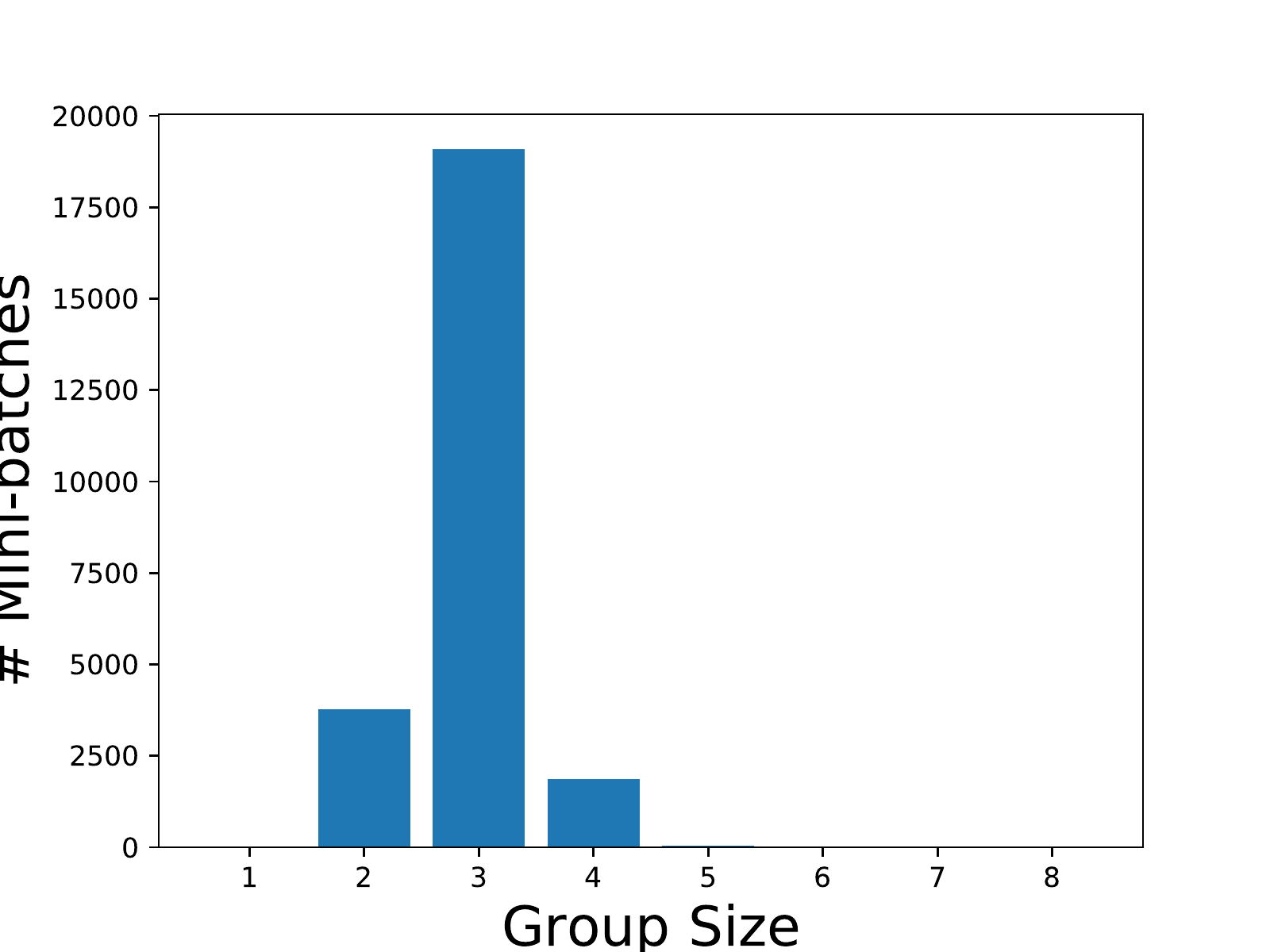}
\subfloat[{FEMNIST}\label{fig:femnist-group-size}]{\hspace{.5\linewidth}}\\
\includegraphics[width=2.0in]{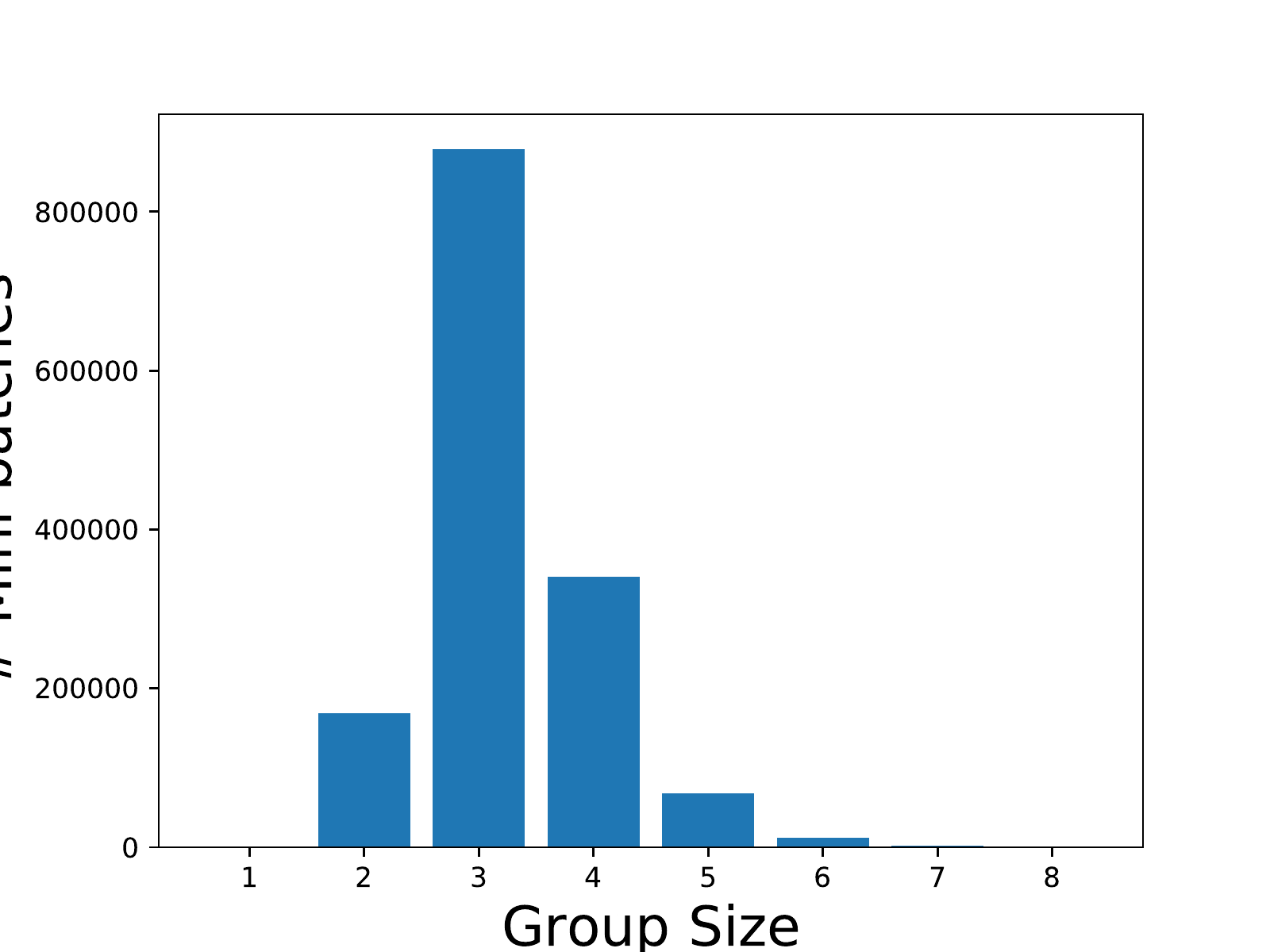}
\subfloat[{Shakespeare}\label{fig:shakespeare-group-size}]{\hspace{.5\linewidth}}
\caption{\footnotesize Number of mini-batches with subject group sizes
  over the entire training run for FEMNIST (a) and Shakespeare (b).
}
\label{fig:group}
\end{figure}

\autoref{fig:accuracy-loss} shows performance of the models trained
using our algorithms.  \fedavg\ performs the best since it does not
incur any DP enforcement penalties.  Item level privacy enforcement
in \localdpsgd\ results in performance degradation of $8\%$ for
FEMNIST and $22\%$ for Shakespeare.  The utility cost of user level
LDP in \userlocaldpsgd\ is quite clear from the figure.  This cost is
also reflected in the relatively high observed loss for the respective
model.  \localgroupdp\ performs significantly better
than \userlocaldpsgd, but worse than \localdpsgd, by $15\%$ on
FEMNIST, and $18\%$ on Shakespeare.  The reason for \localgroupdp{}'s
worse performance is clear from~\autoref{fig:group}(a) and
(c): the group size for a mini-batch tends to be dominated by $3$ on
both FEMNIST and Shakespeare, which cuts the privacy budget for these
mini-batches by a factor of $3$, leading to greater Gaussian noise,
which in turn leads to model performance degradation.

\higradavg\ performs worse than \localgroupdp\ because we need to use
the sampling fraction of largest cardinality subjects at federation
users when calculating mini-batch noise scale
(\autoref{thm-hi-grad-avg-dp}).  This amounts to noise scale
amplification by approximately an order of magnitude.  This
amplification is much higher for Shakespeare (with an average of
$4,484$ data items per subject) by another order of magnitude, because
of which \higradavg{}'s model's utility is the worst.  Moreover,
privacy loss amplification due to horizontal composition limits the
amount of training thereby further limiting model utility (in all our
algorithms).

\subsection{Effect of Subject Data Distribution}

While evaluation of our algorithms using a uniform distribution of
subject data among federation users is a good starting point, often
times the data distribution is non-uniform in real world settings.  To
emulate varying subject data distributions, we conduct experiments on
the FEMNIST dataset where subject data is distributed among federation
users according to the power distribution
\begin{align*}
P(x;\alpha) = \alpha x^{\alpha - 1}, 0 \le x \le 1, \alpha > 0
\end{align*}

\begin{figure}
\centering
\includegraphics[width=2.0in]{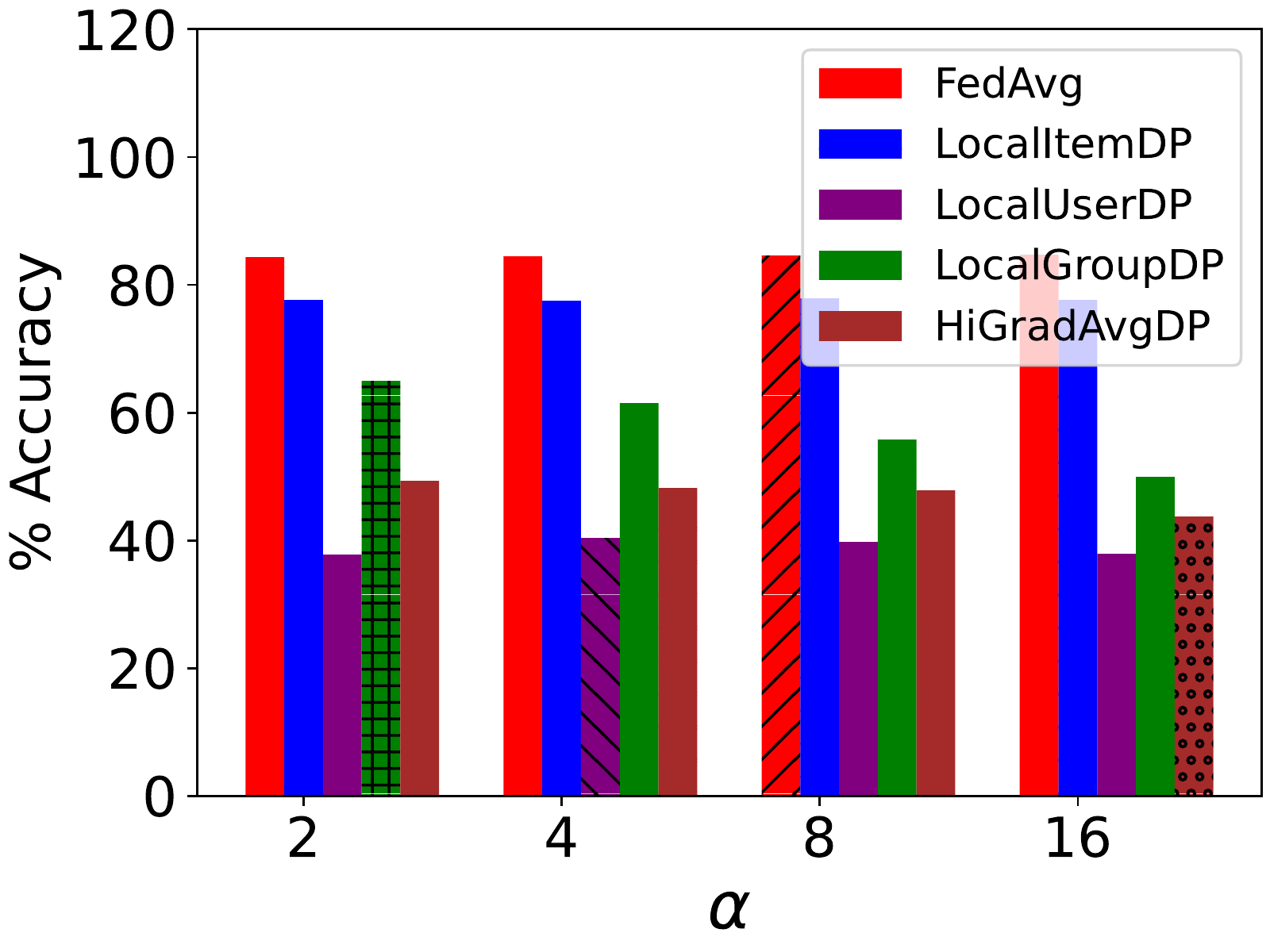}
\caption{\footnotesize Model performance over FEMNIST
dataset of our algorithms over different subject data distributions
dictated by the parameter $\alpha$ of the power distribution.
}
\label{fig:femnist-power-dist}
\end{figure}

\autoref{fig:femnist-power-dist} shows performance of the models
trained using our algorithms over varying subject data distributions
of FEMNIST.  As expected, different data distributions clearly do not
significantly affect \fedavg, \localdpsgd, and \userlocalsgd.
However, performance of the model trained using
\localgroupdp\ degrades noticeably as the unevenness of data
distribution increases, resulting in test accuracy under $50\%$ for
$\alpha = 16$.  This degradation is singularly attributable to growth
in subject group size per mini-batch -- the average group size per
mini-batch ranges from $3$ when $\alpha = 2$ to $6$ when $\alpha =
16$.  This increase in group size significantly reduces the privacy
budget leading to increase in Gaussian noise that restricts test
accuracy.  On the other hand, though \higradavg{}'s model utility is
much lower than that of \localgroupdp{}'s, it appears to be much more
resilient to non-uniform subject data distributions among federation
users up to $\alpha = 8$, and therafter drops noticeably at $\alpha =
16$.

\section{Conclusion}
\label{sec:conclusion}

While various prior works on privacy in FL have explored DP guarantees
at the user and item levels~\cite{liu20,mcmahan18}, to the best of our
knowledge, no prior work has studied subject level granularity for
privacy in the FL setting.
In this paper, we presented a formal definition of subject level DP.
We also presented three novel FL training algorithms that guarantee
subject level DP by either enforcing user level LDP (\userlocaldpsgd),
local group DP (\localgroupdp), or by applying hierarchical gradient
averaging to obfuscate a subject's contribution to mini-batch
gradients (\higradavg).  Our formal analysis over convex loss
functions shows that all our algorithms affect utility loss in
interesting ways, with \localgroupdp\ incurring lower utility loss
than \userlocaldpsgd\ and \higradavg.  Our empirical evaluation on the
FEMNIST and Shakespeare datasets aligns with our formal analysis
showing that while both \userlocaldpsgd\ and \higradavg\ can
significantly degrade model performance, \localgroupdp\ tends to incur
much less loss in model performance compared to \localdpsgd, an
algorithm that provides a weaker item level privacy guarantee.  We
also observe an interesting new aspect of \emph{horizontal
composition} of privacy loss for subject level privacy in FL that
results in model performance degradation.  Both our formal and
empirical analysis demonstrate that there remains significant room for
model utility improvements in algorithms that guarantee subject level
DP in FL.

\bibliographystyle{authordate1}
\bibliography{main}


\appendix

\section{Additional Proofs}
\label{sec:additional-proofs}

The following theorem is a restatement of~\autoref{thm:group-dp} and,
in our notation, of Theorem $2.2$ and the associated footnote $1$ from
Dwork and Roth~\cite[\S2.3]{dwork14}:

\begin{theorem}
Any ($\varepsilon,\delta$)-differentially private randomized algorithm
$\mathcal{A}:\mathcal{D} \rightarrow \mathcal{R}$ is
($k \varepsilon,k e^{(k-1)\varepsilon}\delta$)-differentially private for groups of size $k \geq 2$.
That is, for all $D, D' \in \mathcal{D}$ such that $D$ and $D'$ differ in at most $k$ data items,
and for all $S \subseteq \mathcal{R}$,
\begin{eqnarray*}
\mathrm{Pr}[\mathcal{A}(D) \in S] & \le & e^{k\varepsilon} \mathrm{Pr}[\mathcal{A}(D') \in S] + {\displaystyle {e^{k\varepsilon}-1  \over e^{\varepsilon}-1}} \delta \\
& \leq & e^{k\varepsilon} \mathrm{Pr}[\mathcal{A}(D') \in S] + k e^{(k-1)\varepsilon} \delta \\
& < & e^{k\varepsilon} \mathrm{Pr}[\mathcal{A}(D') \in S] + k e^{k\varepsilon} \delta \\
\end{eqnarray*}
\label{app-thm:group-dp}
\end{theorem}

\begin{proof}
 Suppose that $D$ and $D'$ differ in exactly $k$ data items. Choose any order for these items and call them $d_1, d_2, \ldots, d_k$.
Let $D_0, D_1, \ldots, D_k$ be $k+1$ datasets such that $D_0=D$, and $D_k=D'$, and for all $1 \leq i \leq k$ it holds that $D_{i-1}$ and
$D_i$ differ by exactly one data item, namely $d_i$.
It follows that:

$$
\begin{array}{lll}
\rlap{$\mathrm{Pr}[\mathcal{A}(D_0) \in S]$\hss} \\
& \leq & e^{\varepsilon} \mathrm{Pr}[\mathcal{A}(D_1) \in S] + \delta \\
& \leq & e^{\varepsilon} (e^{\varepsilon}\mathrm{Pr}[\mathcal{A}(D_2) \in S] + \delta) + \delta \\
& \leq & e^{\varepsilon} (e^{\varepsilon} (e^{\varepsilon}\mathrm{Pr}[\mathcal{A}(D_3) \in S] + \delta) + \delta) + \delta \\
& \leq & \cdots \\
& \leq & e^{\varepsilon} (e^{\varepsilon} (e^{\varepsilon} \cdots (e^{\varepsilon}\mathrm{Pr}[\mathcal{A}(D_k) \in S] + \delta) \cdots + \delta) + \delta) + \delta \\
& \leq & e^{k\varepsilon} \mathrm{Pr}[\mathcal{A}(D') \in S] + {\displaystyle \left( \sum_{0\leq i \leq k-1} e^{i\varepsilon} \right)} \delta \\
& = & e^{k\varepsilon} \mathrm{Pr}[\mathcal{A}(D') \in S] + {\displaystyle {e^{k\varepsilon}-1  \over e^{\varepsilon}-1}} \delta 
\end{array}
$$

\noindent
This is the tightest possible bound using this proof technique (also
presented as Lemma $1$ in~\cite{jagielski20}). However, the expression
${e^{k\varepsilon}-1 \over e^{\varepsilon}-1}\delta$ can be unwieldy
for some purposes. Other authors~\cite{dwork14,vadhan17} state this
theorem with a slightly larger value than necessary, perhaps for the
sake of conciseness: because $e^{i\varepsilon} \leq
e^{(k-1)\varepsilon}$ for any $i \leq k-1$, $\left(\sum_{0\leq i \leq
  k-1} e^{i\varepsilon}\right)\delta \leq \left(\sum_{0\leq i \leq
  k-1} e^{(k-1)\varepsilon}\right)\delta = k
e^{(k-1)\varepsilon}\delta$~\cite{dwork14} (equality with
${e^{k\varepsilon}-1 \over e^{\varepsilon}-1}\delta$ holds only when
$k=1$).  Sometimes $k e^{(k-1)\varepsilon}\delta$ is further
simplified to $k e^{k\varepsilon}\delta$~\cite{vadhan17}, which is
strictly larger for all $k \geq 1$.

Because all four of the expressions $e^{k\varepsilon}$,
${e^{k\varepsilon}-1 \over e^{\varepsilon}-1}\delta$, $k
e^{(k-1)\varepsilon}\delta$, and $k e^{k\varepsilon}\delta$ decrease
(strictly) monotonically as $k$ decreases, the same statements apply
even if $D$ and $D'$ differ in fewer than $k$ data items.  Thus one
can say that $\mathcal{A}$ is either ($k \varepsilon,
{e^{k\varepsilon}-1 \over e^{\varepsilon}-1}\delta$)-differentially
private, ($k \varepsilon, k
e^{(k-1)\varepsilon}\delta$)-differentially private, or even ($k
\varepsilon, k e^{k\varepsilon}\delta$)-differentially private for
groups of size $k$; the first of these three statements is the
tightest bound on the failure probability. 
\end{proof}

\begin{theorem}[Same as \autoref{thm-user-local-dp-sgd}]
  \userlocaldpsgd\ with parameter updates
  satisfying~\autoref{eq:randomized-response} as observed by the
  federation server in a training round, enforces user level
  ($\varepsilon$,$\delta$)-local differential privacy provided the
  noise parameter $\sigma$ satisfies the inequality
  from~\autoref{thm:sigma-for-userldp}, and the following inequality
  \begin{align*}
    \sigma > \frac{1}{\sqrt{2 \pi} \varepsilon \delta e^{\varepsilon}}
  \end{align*}
\label{app:thm-user-local-dp-sgd}
\end{theorem}
\begin{proof}
  Let $O(u_1)$ and $O(u_2)$ be the norms of the unperturbed outputs of
  a training round for users $u_1$ and $u_2$ respectively.

  Assuming w.l.o.g. training for $T$ mini-batches per training round
  at $u_1$ and $u_2$, from~\sref{Lemma}{lemma-user-local-dp-sgd} we
  get
  \begin{align*}
    O(u_1) \le \eta T C\\
    O(u_2) \le \eta T C
  \end{align*}

  Now the privacy loss random variable of interest is
  \begin{align*}
    \bigg\vert log \frac{\mathcal{P}[F_l(D_{u_1}] =
      s}{\mathcal{P}[F_l(D_{u_2})] = s} \bigg\vert = \bigg\rvert log
    \frac{exp(\frac{-1}{2 \sigma^2} O(u_1)^2)}{exp(\frac{-1}{2
        \sigma^2} O(u_2)^2)} \bigg\rvert\\
    = \bigg\rvert log\ exp(\frac{-1}{2 \sigma^2} (O(u_2)^2 -
    O(u_1)^2)) \bigg\rvert\\
    = \frac{1}{2 \sigma^2} |O(u_2)^2 - O(u_1)^2|
  \end{align*}

  This quantity is upper bounded by $\varepsilon$ if
  \begin{align*}
    |O(u_2)^2 - O(u_1)^2| \le 2 \varepsilon \sigma^2
  \end{align*}

  For the failure probability bound $\delta$ we need
  \begin{align*}
    \mathcal{P}[|O(u_2)^2 - O(u_1)^2| \ge 2 \varepsilon \sigma^2] <
    \delta
  \end{align*}

  We use the tail bound for Gaussian distributions
  \begin{align*}
    \mathcal{P}[x > t] \le \frac{\sigma}{\sqrt{2 \pi}} \frac{1}{t}
    exp(\frac{-t^2}{2 \sigma^2})
  \end{align*}

  Because we are concerned with just $|O(u_2)^2 - O(u_1)^2|$ we will
  find $\sigma$ such that
  \begin{align*}
    \frac{\sigma}{\sqrt{2 \pi}} \frac{1}{t} exp(\frac{-t^2}{2
      \sigma^2}) < \frac{\delta}{2}
  \end{align*}

  Substituting $t = 2 \varepsilon \sigma^2$ we get
  \begin{align*}
    \frac{\sigma}{\sqrt{2 \pi}} \frac{1}{2 \varepsilon \sigma^2}
    exp(\frac{-2 \varepsilon \sigma^2}{2 \sigma^2}) < \frac{\delta}{2}
  \end{align*}
  which resolves to
  \begin{align*}
    \sigma > \frac{1}{\sqrt{2 \pi} \varepsilon \delta e^{\varepsilon}}
  \end{align*}

\end{proof}

\begin{theorem}[Same as \autoref{thm-horizontal-composition}]
  Given a FL training algorithm
  $\mathcal{F}=(\mathcal{F}_l,\mathcal{F}_g)$, in the most general
  case where a subject's data resides in the private datasets of
  multiple federation users $u_i$, the aggregation algorithm
  $\mathcal{F}_g$ adaptively composes subject level privacy losses
  incurred by $\mathcal{F}_l$ at each federation user.
  \label{app:thm-horizontal-composition}
\end{theorem}
\begin{proof}
  Assume two distinct users $u_1$ and $u_2$ in a federation that host
  private data items of subject $s$.  Let $\varepsilon_1$ and
  $\varepsilon_2$ be the respective subject privacy losses incurred by
  the two users during a training round.

  It is straightforward to see that, in the worst case, data items of
  $s$ at users $u_1$ and $u_2$ can affect disjoint parameters in
  $\mathcal{M}$.  Thus parameter averaging done by $\mathcal{F}_g$
  simply results in summation and scaling of these disjoint parameter
  updates.  As a result, the privacy losses, $\varepsilon_1$ and
  $\varepsilon_2$ incurred by $u_1$ and $u_2$ respectively are
  retained to their entirety by $\mathcal{F}_g$.  In other words,
  privacy losses incurred for subject $s$ at users $u_1$ and $u_2$
  compose adaptively.
\end{proof}

\begin{theorem}[Same as \autoref{thm-horizontal-composition-rounds}]
  Consider a FL training algorithm
  $\mathcal{F}=(\mathcal{F}_l,\mathcal{F}_g)$ that samples $s$ users
  per training round, and trains the model $\mathcal{M}$ for $R$
  rounds.  Let $\mathcal{F}_l$ at each participating user, over the
  aggregate of $R$ training rounds, locally enforce subject level
  ($\varepsilon$,$\delta$)-DP.  Then $\mathcal{F}$ globally enforces
  the same subject level ($\varepsilon$,$\delta$)-DP guarantee by
  training for $\frac{R}{\sqrt{s}}$ rounds.
  \label{app:thm-horizontal-composition-rounds}
\end{theorem}
\begin{proof}
  The proof of training round constraints on horizontal composition
  can be broken down into two cases: First, each user in the
  federation locally trains for exactly $T$ mini-batches per training
  round, with exactly the same mini-batch sampling probability $q$.
  Since horizontal composition is equivalent to adaptive composition
  in the worst case, the moments accountant method shows us that the
  resulting algorithm will be $(O(q \varepsilon \sqrt{TRs}),
  \delta)$-differentially private.  To compensate for the $\sqrt{s}$
  factor scaling of the privacy loss, $\mathcal{F}$ can be executed
  for $\frac{R}{\sqrt{s}}$ training rounds, yielding a $(O(q
  \varepsilon \sqrt{TR}), \delta)$-differentially private algorithm.

  In the second case, each user $u_i$ may train for a unique number of
  mini-batches per training round, with a unique mini-batch sampling
  probability dictated by $u_i$'s private dataset.  Let $T_1, T_2,
  ..., T_s$, and $q_1, q_2, ..., q_s$ be the number of mini-batches
  per training round and mini-batch sampling fraction for the sampled
  users $u_1, u_2, ..., u_s$ respectively.

  All our algorithms \ref{algo-user-local-dp-sgd},
  \ref{algo:local-group-dp}, and \ref{algo:hi-grad-avg-dp} locally
  enforce subject level ($O(q_i \varepsilon \sqrt{T_i R}
  )$,$\delta$)-DP at each user $u_i$.  Privacy enforcement is done
  independently at each federation user $u_i$.  Furthermore, note that
  the privacy loss is uniformly apportioned among training rounds.
  Let $\mathcal{E} = O(q_i \varepsilon \sqrt{T_i R})$.  Note that
  $\mathcal{E}$ is identical for each user $u_i$ in the
  federation. Thus if $\mathcal{E}$ is the total privacy loss budget
  over $R$ training rounds, a sampled user incurs $\varepsilon_r =
  \mathcal{E}/R$ privacy loss in a single training round $r$.
  Similarly, each of the $s$ sampled users in round $r$ incurs
  identical privacy loss $\varepsilon_r$ despite having different
  mini-batches per training round $T_i$ and mini-batch sampling
  probabilities $q_i$s.  As noted earlier, these privacy losses
  compose horizontally (adaptively) via $\mathcal{F}_g$ over $s$
  users, leading to privacy loss amplification by a factor of
  $\sqrt{s}$ as per the moments accountant method.  Given a fixed
  privacy loss budget $\mathcal{E}$, to compensate for this privacy
  loss amplification, $\mathcal{F}$ can be executed for
  $\frac{R}{\sqrt{s}}$ training rounds.
\end{proof}

\end{document}